\documentclass{ieeeaccess}
\usepackage{cite}
\usepackage{amsmath,amssymb,amsfonts}
\usepackage{algorithm}
\usepackage{algorithmicx}
\usepackage{algpseudocode}
\usepackage{graphicx}
\usepackage{textcomp}
\usepackage{lipsum}
\usepackage{threeparttable}
\usepackage{booktabs}  % For well-spaced lines
\usepackage{multirow}
\usepackage{array}
\usepackage{graphicx} % for \checkmark symbol
\usepackage{lscape}  % optional, for wide tables
\usepackage{booktabs}  % For well-spaced lines
\usepackage{xcolor}
\usepackage{etoolbox} % Add this to your preamble if not already included

\usepackage{bm}
\makeatletter
\AtBeginDocument{\DeclareMathVersion{bold}
\SetSymbolFont{operators}{bold}{T1}{times}{b}{n}
\SetSymbolFont{NewLetters}{bold}{T1}{times}{b}{it}
\SetMathAlphabet{\mathrm}{bold}{T1}{times}{b}{n}
\SetMathAlphabet{\mathit}{bold}{T1}{times}{b}{it}
\SetMathAlphabet{\mathbf}{bold}{T1}{times}{b}{n}
\SetMathAlphabet{\mathtt}{bold}{OT1}{pcr}{b}{n}
\SetSymbolFont{symbols}{bold}{OMS}{cmsy}{b}{n}
\renewcommand\boldmath{\@nomath\boldmath\mathversion{bold}}}
\makeatother

\algrenewcommand\algorithmicensure{\textbf{Returns:}}

\def\BibTeX{{\rm B\kern-.05em{\sc i\kern-.025em b}\kern-.08em
    T\kern-.1667em\lower.7ex\hbox{E}\kern-.125emX}}

\usepackage{atbegshi}

\newcommand\submittedtext{%
  \footnotesize This work has been submitted to the IEEE for possible publication. Copyright may be transferred without notice, after which this version may no longer be accessible.}

\AtBeginShipout{%
  \AtBeginShipoutUpperLeft{%
    \raisebox{-2\baselineskip}{%
      \parbox{\dimexpr\textwidth-1.5cm}{\centering\color{red}\fbox{\parbox{\dimexpr0.65\textwidth-\fboxsep-\fboxrule\relax}{\submittedtext}}}%
    }%
  }%
}

\begin{document}
\history{Date of publication xxxx 00, 0000, date of current version xxxx 00, 0000.}
\doi{10.1109/ACCESS.2024.0429000}

\title{Enhancing Predictive Maintenance in Mining Mobile Machinery through a TinyML-enabled Hierarchical Inference Network}
\author{\uppercase{Raúl de la Fuente}\authorrefmark{1},
\uppercase{Luciano Radrigan}\authorrefmark{2}, \uppercase{Anibal S Morales}\authorrefmark{3}}

\address[1]{Computer Science Department, Universidad de Chile, Beauchef 851, Santiago 8370456, Chile (e-mail: raul.delafuente@ing.uchile.cl)}
\address[2]{Electrical Engineering, Department, Universidad de Concepción, Concepción 4030000, Chile (e-mail: lradrigan@udec.cl)}
\address[3]{Centro de Transicion Energetica, Facultad de Ingenieria, Arquitectura y Diseño (FIAD), Universidad San Sebastian, Concepcion 4030000, Chile (e-mail: anibal.moralesm@uss.cl)}

\markboth
{R. de la Fuente \headeretal: ESN-PdM: Enhancing Condition Monitoring in Non-Stationary Mining Machinery}
{R. de la Fuente \headeretal: ESN-PdM: Enhancing Condition Monitoring in Non-Stationary Mining Machinery}

\corresp{Corresponding author: Anibal S Morales (e-mail: anibal.moralesm@uss.cl).}

\begin{abstract}
Mining machinery operating in variable environments faces high wear and unpredictable stress, challenging Predictive Maintenance (PdM). This paper introduces the Edge Sensor Network for Predictive Maintenance (ESN-PdM), a hierarchical inference framework across edge devices, gateways, and cloud services for real-time condition monitoring. The system dynamically adjusts inference locations—on-device, on-gateway, or on-cloud—based on trade-offs among accuracy, latency, and battery life, leveraging Tiny Machine Learning (TinyML) techniques for model optimization on resource-constrained devices. Performance evaluations showed that on-sensor and on-gateway inference modes achieved over 90\% classification accuracy, while cloud-based inference reached 99\%. On-sensor inference reduced power consumption by approximately 44\%, enabling up to 104 hours of operation. Latency was lowest for on-device inference (3.33 ms), increasing when offloading to the gateway (146.67 ms) or cloud (641.71 ms). The ESN-PdM framework provides a scalable, adaptive solution for reliable anomaly detection and PdM, crucial for maintaining machinery uptime in remote environments. By balancing accuracy, latency, and energy consumption, this approach advances PdM frameworks for industrial applications.
\end{abstract}

\begin{keywords}
IoT, ESP32, edge computing, cloud computing, deep learning, machine learning, TensorFlow, TinyML, heavy machinery, predictive maintenance, minning.
\end{keywords}

\titlepgskip=-21pt

\maketitle

\section{Introduction}
\label{sec:introduction}
\PARstart{T}{HE} mining sector is a vital component of the global resource economy, providing the raw materials essential for industrial production and infrastructure development. By 2026, the industry is projected to reach a market value of \$3.36 trillion  \cite{mining-industry-market-report-2024}, driven largely by the growing demand for minerals such as lithium, cobalt, and copper, which are critical components in renewable energy technologies  \cite{mining-industry-renewable-energy}, electric vehicles \cite{mining-industry-electric-vehicles}, and consumer electronics \cite{mining-industry-consumer-electronics}. Mining operations are inherently complex, encompassing multiple stages like exploration, extraction, processing, and transportation. To maintain efficiency and safety, the sector greatly depends on a wide range of machinery and mobile and semi-mobile equipment, including drilling rigs, shovels, excavators, haul trucks, front-loaders, and other auxiliary equipment \cite{mining-industry-heavy-machinery}. These operations often occur in harsh, remote environments, exposing assets to extreme conditions such as high temperatures, humidity, dust, and heavy loads \cite{mining-industry-maintenance}. Prolonged exposure to these conditions leads to equipment degradation, reduced remaining useful life, increased maintenance costs, and elevated safety risks.

PdM has become critical for optimizing mining operations, offering benefits such as improved system availability, cost savings, and enhanced failure prediction \cite{mining-industry-pdm}.By leveraging data from continuous condition monitoring sensors, PdM enables proactive decision-making and timely maintenance, reducing the risk of unplanned downtime \cite{pdm-1}. The rise of Artificial Intelligence (AI) has further advanced PdM, as Machine Learning (ML) and Deep Learning (DL) algorithms can analyze vast datasets, identify patterns, and predict equipment failures more accurately than traditional methods  \cite{pdm-2}. Internet of Things (IoT) technologies, particularly Wireless Sensor Networks (WSNs), have become key components for collecting real-time data on machinery performance \cite{related-work-cloud-inf-2,related-work-gateway-inf-2, related-work-gateway-inf-3}. WSNs consist of spatially distributed sensor nodes and gateways that communicate wirelessly to collect and transmit data \cite{wsn-1}. Typically, these sensor nodes are small-size, light-weight, energy-efficient, cost-effective and remarkably flexible to deploy, making them ideal for mining environments \cite{mining-industry-wsn}.

PdM frameworks are structured methodologies that encompass the entire PdM process: from data collection, preprocessing, and communication to ML and DL model development, training, and deployment. Traditional PdM frameworks often rely on a fixed inference location, either at the cloud in a dedicated serverless service (in a server on-premise far away from the operation instead) or closer to the edge on a gateway or sensor node \cite{related-work-cloud-inf-1,related-work-gateway-inf-1,related-work-sensor-inf-1}. Each approach has its advantages and limitations. Cloud-based inference offers superior accuracy and scalability at the cost of high latency and the need for stable network connectivity  \cite{related-work-cloud-inf-2,related-work-cloud-inf-3}. Edge-based inference, both on gateways and nodes, minimizes latency and enables real-time decision-making but may increase power consumption and limited model complexity \cite{related-work-gateway-inf-2,related-work-gateway-inf-3,related-work-sensor-inf-2,related-work-sensor-inf-3}. 

\textit{Motivation:} Each condition monitoring approach and PdM solution has its own strengths and weaknesses. Regardless of the inference approach, there is not a single technique that can detect, diagnose and predict all types of faults optimally. This call for more flexible and adaptive solutions for faster and accurate maintenance predictions under uncertainty in operational conditions. In this context, the inference location is critical to ensure high the system’s performance. The inference location in a hierarchical inference network with several levels directly determinates both the speed and accuracy of detected events. A fixed inference location may be inadequate due to dynamic changes in operational conditions over time. For instance, different expertise of machinery operators or mining fronts with different shape and leveling, expose equipment to different stress and wear levels. By leveraging on-cloud, on-gateway, and on-device inference capabilities, the system dynamically can adjusts inference locations based on trade-offs between real-time demands and conditions such as accuracy, latency, and battery range. By adapting the inference location dynamically, the PdM can be optimized the condition monitoring process, leveraging cloud resources when accuracy is critical and shifting to edge computing when real-time decision-making is mandatory.

This paper presents a PdM framework that integrates edge inference approaches (such as on-gateway and on-sensor-based inference) and cloud computing services into an unified hierarchical inference system to enhance real-time condition monitoring of heavy machinery. The proposed framework leverages the strengths of each inference level to provide real-time and energy-efficient condition monitoring, adapting the inference location based on operational demands and conditions such as latency, accuracy, and energy consumption. The ESN-PdM system is evaluated through a case-study in a real-world industrial scenario, where vibration data from mining equipment is used to evaluate the operational state of the machinery and triggering alarms when anomalies arise. The main contributions of this work are as follows:

\begin{itemize}
    \item An open-source, end-to-end framework for condition monitoring and PdM of mobile mining machinery in non-stationary operations.
    \item A novel adaptive inference mechanism that dynamically updates the inference location for a node based on operational conditions.
    \item A guide on how to use state-of-the-art TinyML optimization approaches to achieve optimal accuracy and model compression for efficient deployment of DL models on limited hardware resources of IoT edge devices.
    \item A comprehensive evaluation of the proposal in terms of operational status classification accuracy, inference latency, and node energy consumption through a real-world industrial case-study.
\end{itemize}

This paper is organized as follows: Section II discusses about the applicability of time-varying classification of multivariate time series from mechanical and/or electrical systems, and addresses the further implementation challenges on TinyML and PdM frameworks. Section III presents the proposed ESN-PdM framework, describing its architecture, components, and adaptive inference mechanisms. Section IV introduces a case study based on DL strategies for PdM. In Section V, the ESN-PdM framework is evaluated in terms of classification accuracy for PdM, inference latency, and energy consumption. Finally, Section VI presents a summary of main findings and conclusions, and outlines future research directions.

\section{Related Work}
\label{sec:related-work}

\subsection{TinyML and Edge Computing}

% TinyML: a response to cloud-based inference
Remote environments like mining sites often lack of reliable network connectivity, making cloud-based inference impractical for real-time applications under some operating conditions. Also, the transmission of raw sensor data to the cloud introduces significant communication overhead in terms of latency, bandwidth, and energy consumption \cite{mining-industry-pdm}. Consequently, efforts have been made on both hardware and software fronts to bring ML and DL capabilities to the edge, allowing these devices to perform intelligent tasks locally without relying on cloud services. TinyML, a field at the intersection of ML and embedded systems, aims to empower constrained hardware devices such as wearables, wireless sensors, smartphones, and microcontroller units (MCUs) with on-device ML capabilities \cite{tinyml-2}. By deploying ML models at the edge, TinyML systems can achieve real-time inference, enhanced privacy, low power consumption, and reduced connectivity requirements \cite{tinyml-3,tinyml-4}.

% Hardware for TinyML
Hardware innovations have been instrumental in enabling TinyML applications. Advancements in MCUs, single board computers (SBCs), field programmable gate arrays (FPGAs), and application specific integrated circuits (ASICs) have been vital in developing platforms suitable for TinyML  \cite{tinyml-1}. Notable examples include the NVIDIA Jetson Nano, a low-cost SBC featuring a GPU for accelerating DL models \cite{nvidia_jetson_nano}; the Google Coral Edge TPU, an ASIC optimized for fast, low-power ML inference \cite{google_coral_edge_tpu}; and AMD's Xilinx Zynq UltraScale+ MPSoC, which combines FPGA flexibility with ARM cores for custom AI accelerator designs \cite{amd_xilinx_zynq_ultrascale_mpsoc}. 

% Software for TinyML
TinyML software focuses on lightweight ML libraries that exploit the full potential of hardware platforms and optimize the deployment of ML models on edge devices. TensorFlow Lite for Microcontrollers (TFLM) is a prime example, providing tools for model conversion, optimization, and deployment on MCUs \cite{tensorflow_lite_microcontrollers}. Other remarkable libraries include ARM’s CMSIS NN, a set of optimized kernels for neural network operations on Cortex M processors \cite{arm_cmsis_nn}, and NVIDIA TensorRT, a high-performance DL inference engine for edge devices \cite{nvidia_tensorrt}.

% TinyML optimization techniques for ANNs
Artificial neural networks (ANNs) are commonly used in TinyML applications due to the native parallelism of DL algorithms, which can be leveraged by modern hardware accelerators \cite{tinyml-7}. There is significant interest in optimizing these models for deployment on resource constrained devices. Common optimization techniques include quantization, pruning, knowledge distillation, and model compression.

\subsubsection{Quantization}

Quantization method reduces the precision of numerical values, typically from 32-bit floating point to 8-bit integers \cite{tinyml-1}. Applying quantization to the parameters of a trained ANN significantly decreases the model’s memory footprint and inference time \cite{tinyml-6}. Less precise data types require less memory and operate faster, reducing memory access and computation overhead. This makes the model more power-efficient, which is especially advantageous for deployment on battery-powered, resource-constrained IoT devices. However, quantization can lead to a loss of precision, potentially affecting model performance. Techniques like Quantization Aware Training (QAT) help mitigate this by simulating quantization during training, allowing the model to adapt to the reduced precision \cite{tinyml-7}. Figure \ref{fig:tinyml-optimization}.a illustrates the quantization process, showing the conversion of 32-bit floating point weights to 8-bit integers.

\subsubsection{Pruning}

Pruning method optimizes ANNs by eliminating redundant components such as weights, neurons, or even entire layers \cite{tinyml-2}. It can be applied either during training or post-training. When applied during training, pruning acts as a regularizer, reducing overfitting and leading to a more compact model as training progresses \cite{tinyml-5}.It can be applied either during training or post-training. When applied during training, pruning acts as a regularizer, reducing overfitting and leading to a more compact model as training progresses \cite{tinyml-1}. Figure \ref{fig:tinyml-optimization}.b illustrates the pruning process, showing the removal of redundant weights from a neural network.
This paper tests Quantization and Pruning methods for evaluations on case-study in next sections. For the sake of completeness, Knowledge distillation and Tensor decomposition are described below. 

\subsubsection{Knowledge Distillation}

Knowledge distillation transfers knowledge from a large, complex model (the teacher) to a smaller, simpler model (the student) \cite{tinyml-5}. The process takes into account two main factors: the type of knowledge and the distillation scheme. Knowledge types include response-based, where the student mimics the teacher’s final predictions; feature-based, where the student learns from the teacher’s intermediate layers; and relation-based, where the student learns from the relation- ships between layers or samples \cite{tinyml-2}. Distillation schemes include offline distillation, where the teacher is trained first and then guides the student; online distillation, where both models train simultaneously; and self-distillation, where the teacher and student share the same architecture \cite{tinyml-8, tinyml-2}. This method creates lightweight models that can maintain or even surpass the performance of larger models. Figure \ref{fig:tinyml-optimization}.c illustrates the knowledge distillation process, showing the transfer of knowledge from a teacher model to a student model.

\subsubsection{Tensor Decomposition}

Finally, Tensor decomposition is an optimization technique that breaks down large tensors, such as weight matrices in ANNs, into smaller, more manageable components  \cite{tinyml-10}. This process reduces both the number of parameters and computational requirements, enhancing model efficiency, particularly in TinyML applications. For example, CP Decomposition factorizes a tensor into a sum of outer products of vectors, while Tucker Decomposition represents data using a core tensor and factor matrices  \cite{tinyml-9}. Another method, Tensor Train Decomposition, transforms a tensor into a sequence of low-rank tensors, significantly reducing model size and computation while maintaining accuracy \cite{tinyml-11}. Figure \ref{fig:tinyml-optimization}.d illustrates a simplified tensor decomposition process, showing the factorization of a weight matrix into smaller components.
%ESNs implemented in this paper are a type of Recurrent Neural Network (RNN) specifically designed to model dynamic time-series data efficiently. Unlike traditional RNNs, which involve complex training of recurrent weights, ESNs rely on a randomly initialized, fixed-weight reservoir, which simplifies the training process and can make them particularly useful for real-time applications, including condition monitoring and PdM of mobile machinery in non-stationary operation.
%ESNs have gained attention for PdM due to their robustness, simplicity, and efficiency in handling sequential data. Their core strength lies in their ability to capture non-linear dependencies in temporal data, making them well-suited for detecting trends and anomalies in machinery performance over time\cite{DeepESN} \cite{IAEdge}. 

\Figure[ht](topskip=0pt, botskip=0pt, midskip=0pt)[width=1.95\columnwidth]{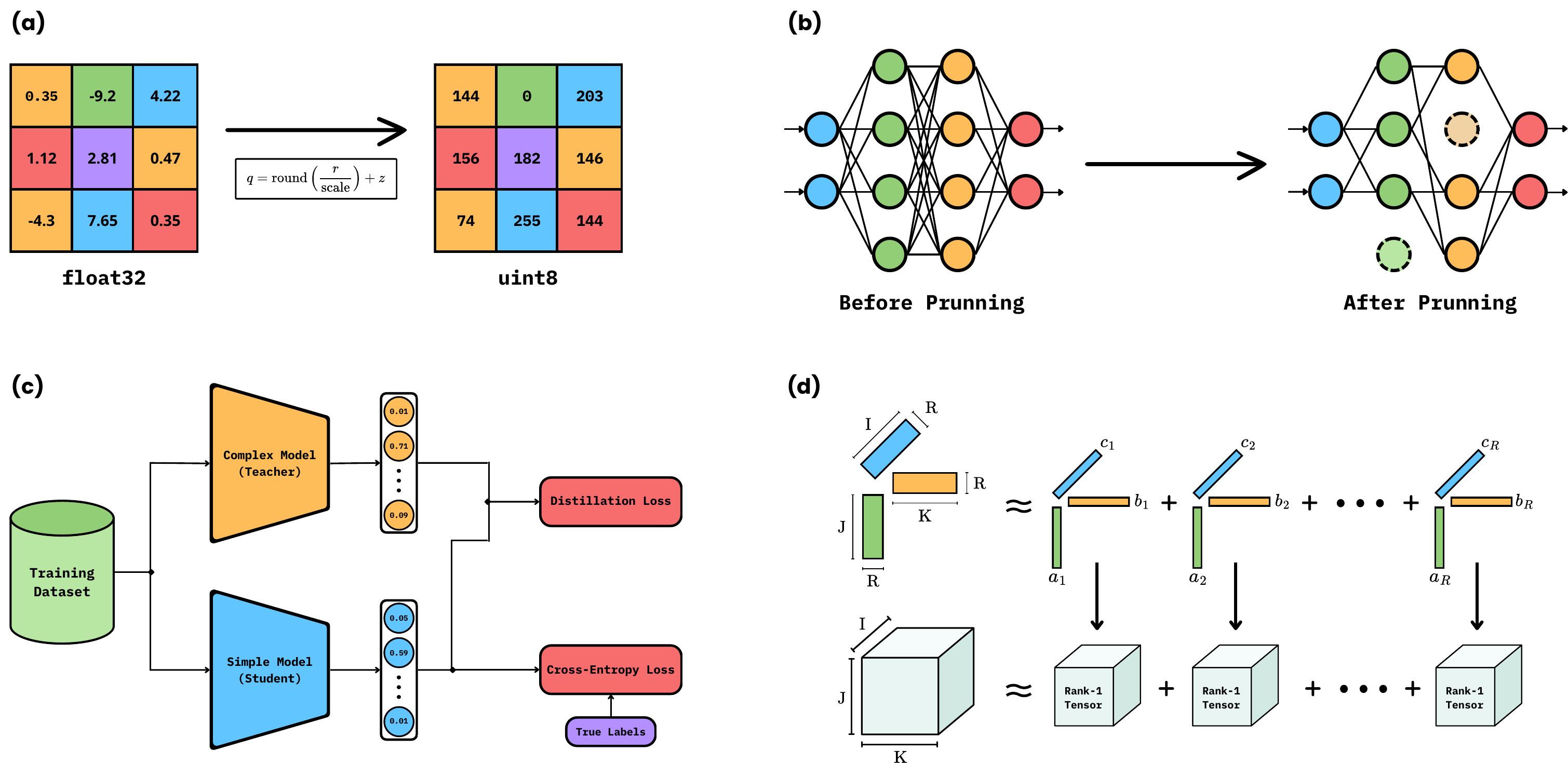}{ \textbf{TinyML optimization techniques: (a) quantization, (b) prunning, (c) knowledge distillation, and (d) tensor decomposition.}\label{fig:tinyml-optimization}}

\subsection{PdM Frameworks}
PdM systems have become essential tools in various industries, enabling proactive maintenance strategies that reduce downtime, maintenance costs, and equipment failures. There is a growing interest in developing data-driven PdM frameworks that integrate state-of-the-art techniques such as WSNs and ML/DL to enhance equipment reliability and operational performance, correspondingly. Within these predictive frameworks, the inference process plays a key role in detecting anomalies, predicting failures, and monitoring equipment health. This section reviews existing work on condition-based PdM, focusing on the inference approaches employed in these frameworks.

\subsubsection{Cloud-based Inference}
% Deep-Learning-Driven Proactive Maintenance Management of IoT-Empowered Smart Toilet
See-To et al. \cite{related-work-cloud-inf-1} presented a condition monitoring system specifically designed for IoT-enabled smart toilets, marking one of the first applications of proactive maintenance strategies in public hygiene facilities. The framework utilized a star-shaped WSN that integrates both wired and wireless sensors to collect real-time data on the operating conditions of toilet equipment. These sensors include cubicle occupancy sensors, odor sensors, people counters, light intensity sensors, soap dispenser sensors, toilet tissue dispenser sensors, and smart energy meters. The collected data was transmitted to an IoT Gateway, which aggregate and send it to a cloud platform for centralized storage and analysis.
The work proposed a DL model designed to predict the operational and environmental conditions of the equipment, combining the strengths of a Convolutional Neural Network (CNN) for feature extraction and a RNN, specifically a Long Short-Term Memory (LSTM) network, for time-series fore- casting. The model was trained on historical sensor data from a public toilet in Hong Kong, and its performance was evaluated using metrics such as root-mean-squared error (RMSE), mean absolute error (MAE), coefficient of variation of RMSE (CV(RMSE)), and R-squared. The proposed model outperformed six traditional ML algorithms. Additionally, a Naïve Bayes classifier was used as an anomaly detector to identify deviations from normal equipment behavior based on the proposed model’s output.

% Development of PdM System for Haemodialysis Reverse Osmosis Water Purification System
Bani et al. \cite{related-work-cloud-inf-2} describe the development of an IoT-based PdM system for the Haemodialysis Reverse Osmosis (RO) Water Purification System. The goal of this study was to ensure the continuous operation of the RO system, essential for haemodialysis treatments, by predicting potential failures and enabling timely maintenance interventions. The system uses an ESP8266 module as a sensor node, integrated with twelve sensors, including conductivity, flow, and pressure sensors, to monitor key parameters of the RO system. Due to the limited number of pins on the ESP8266, a 16-channel analog multiplexer was employed to manage multiple sensor inputs. The module utilizes WiFi connectivity to transmit collected data to a cloud server for storage and inference. An LSTM network is employed to predict equipment breakdowns by detecting anomalies in sensor readings. The model was trained on historical data from an RO system prototype developed for the study. Although the article lacks details on the evaluation metrics used to assess the model's performance, the LSTM model successfully detected anomalies and predicted equipment failures. These predictions were displayed on a monitoring dashboard, enabling authorized personnel to take timely corrective actions. The dashboard interface visualizes real-time data from pressure, conductivity, and flow rate sensors, facilitating easy interpretation and informed decision-making by the users.

% PdM Method using Machine Learning for IoT Connected Computed Tomography Scan Machine
Shah et al. \cite{related-work-cloud-inf-3} proposed a PdM framework for IoT- connected Computed Tomography (CT) scan machines. The framework aims to predict the probability of machine breakdowns by monitoring environmental and operational parameters in real-time. To achieve this goal, the study employed a star-like IoT network comprising various sensors installed on the CT scan machine. Among the sensors employed was a 3-in-1 AM103 Temperature, Humidity and CO2 Sensor, RAD-S101 Radiation Dosimeter, MK926 MEMS accelerometer, and Schneider PM5110 Power Meter. The sensors communicated their measurements to an IoT Gateway, which aggregates the data and transmits it to the cloud over 4G LTE connectivity. The cloud services leveraged by the framework include a database server for data storage and preprocessing, and a dedicated server for both ML model handling and dashboard visualization. For inference, the framework employed Multi-Layer Perceptron (MLP) trained on historical sensor readings collected through the IoT network. The model achieved a high prediction accuracy above 95\% on the test subset, demonstrating its effectiveness in predicting machine breakdowns. Additionally, a confusion matrix analysis was conducted to evaluate the model’s performance, along with an operational effectiveness assessment through real-time monitoring on the dashboard.

\subsubsection{Gateway-based Inference}
% Edge Computing-Assisted IoT Framework With an Autoencoder for Fault Detection in Manufacturing PdM
Yu et al. \cite{related-work-gateway-inf-1} developed a PdM framework for health monitoring of a syngas reciprocating compressor in a remote industrial plant. The framework integrated edge computing and IoT technologies to enable real-time fault detection using an autoencoder model. The system was structured into three layers: edge, cloud, and application. In the edge layer, various industrial sensors were deployed to measure parameters such as vibrations, pressure, temperature, velocity, and humidity. These sensor readings were transmitted to an Open Platform Communications (OPC) server for real-time data collection, while a metadata server logged measurement information.
By aligning sensor metadata with the OPC server readings, the system aggregated real-time data into text files. Then, these files were published to an Apache Kafka cluster for message brokering. The cloud layer handled centralized data integration, storage, and analytics using Amazon Web Services (AWS) and a distributed Hadoop cluster for scalability and fault tolerance. Data from the edge layer was securely transferred via Secure File Transfer Protocol (SFTP) and stored in a data lake. Apache Spark was employed for global optimization and processing of the data.
In the application layer, real-time monitoring, reporting, and interactive analytics were provided through Qlik software. This layer offers user-friendly dashboards and triggers alerts for potential faults, simplifying timely interventions to support proactive maintenance decisions. For fault detection, the Apache Kafka cluster forwarded data to an Apache Spark Streaming cluster, which performed data preprocessing and applied a Distributed Stack Sparse Autoencoder (DSSAE) model. The DSSAE combined multiple autoencoders with sparsity constraints in a stacked architecture, enabling efficient handling of large-scale, high- dimensional datasets through parallel processing across multiple nodes in the Spark cluster. Compared to traditional methods like Principal Component Analysis (PCA), the DSSAE demonstrated superior performance by reducing false alarms and detecting faults with several hours or days of forecast horizon.

% Improving an IoT-Based Motor Health PdM System Through Edge-Cloud Computing
Lee et al. \cite{related-work-gateway-inf-2} proposed a PdM framework that combines edge and cloud computing to enhance an IoT-based health monitoring system for motors. The framework aimed to predict faults and prevent breakdowns by analyzing acoustic signals collected from the motor through Embedded Acoustic Recognition Sensors (EARS), a sensing node developed specifically for the study. The raw data captured by the EARS was transmitted to a preprocessing server via HTTP, where Fast Fourier Trans- form (FFT) was applied to extract frequency components. The, the preprocessed data was normalized and sent to an inference server, which reshapes the data and classifies it into one of nine fault classes using a CNN. Both the preprocessed data and the results were stored in a database for further data analysis and reporting. The study evaluated two setups: a pure cloud setup where all preprocessing, inference and storage tasks were performed in AWS platforms, and a hybrid edge-cloud setup where only storage was handled in the cloud while preprocessing and inference were executed in an edge device.
The edge-enabled PdM system consisted on an star-like IoT network where EARS were connected to an Advantech ICR- 3231 cellular router which served as a gateway to the edge device. The study compared the performance of the pure cloud and edge-cloud setups based on timing breakdown, CPU and memory usage, and overall system performance. The results shown that the edge-cloud setup outperforms the pure cloud setup in terms of execution time and resource usage, highlighting the benefits of edge computing in PdM applications.

% Machine Learning based Real Time PdM at the Edge for Manufacturing Systems: A Practical Example
Ringler et al. \cite{related-work-gateway-inf-3} presented an edge-enabled PdM framework for manufacturing systems, focusing on real-time monitoring and fault diagnosis on an industrial oil-injection screw compressor using a ML-based predictive model. The framework consisted on a star-like WSN, with vibration sensors attached to the compressor’s motor and screw, collecting data at a high sampling frequency of 10 kHz. A Raspberry Pi 4 hosted an MQTT broker for communication between devices, connected via a router and VPN for secure data transmission. Sensor readings were published to the broker and forwarded to edge devices for real-time processing.
The framework employed two NVIDIA Jetson Nano Developer Kits for edge computing, chosen for their low power consumption and high computing capacity. These devices preprocessed the data, extract features, and perform ML inference using various supervised learning models, including ANNs, Decision Trees, Random Forest, KNN, SVM, and Naïve Bayes. The ANN model demonstrated the best diagnostic performance, achieving high accuracy rates with minimal execution time in real- time conditions. Finally, the edge devices sent the results to an InfluxDB instance for storage and visualization on Grafana dashboards.

\subsubsection{Sensor-based Inference}

% Deep-Reinforcement-Learning-Based PdM Model for Effective Resource Management in Industrial IoT
Ong et al. \cite{related-work-sensor-inf-1} presented a deep reinforcement learning based PdM framework designed for effective resource management in Industrial Internet of Things (IIoT) networks. Their approach addresses unplanned equipment breakdowns by leveraging automatically learned decision policies within a stochastic environment. The framework was structured into three levels: the equipment level, the edge cloud level, and the data aggregation and analysis level. At the equipment level, in situ edge sensors are deployed to monitor real-time parameters of industrial equipment, such as temperature, pressure, and vibration. These intelligent devices possess computational capabilities for on-site data processing and anomaly detection using a PdM model.
Upon detecting anomalies, the edge sensors generate metadata that includes probabilistic alarm information and maintenance action recommendations. This metadata is transmitted through the network level, using high-speed communication technologies like Gigabit Ethernet, WiFi (IEEE 802.11ax), and 5G. The network infrastructure comprises routers, switches, and gateways that facilitate data transmission among devices. The recipient of the metadata is the Edge Cloud (EC), an on-premises cloud server equipped with high processing power and memory for big data storage and ML analytics. The EC performs further analysis and decision-making based on the measured data. 
The PdM framework was evaluated using a case study of health monitoring and resource management in a generic production facility. The authors employed a Proximal Policy Optimization LSTM (PPO-LSTM) model, which efficiently learned the optimal decision policy for resource management. Empirical results demonstrate that the PPO- LSTM model outperformed comparable deep reinforcement learning methods in terms of convergence efficiency, simulation performance, and flexibility.

% LOPdM: A Low-Power On-Device PdM System Based on Self-Powered Sensing and TinyML
Chen et al. \cite{related-work-sensor-inf-2} introduced LOPdM, a low-power on-device PdM framework that utilizes self-powered sensors (SPS) and TinyML techniques to fulfill the demands for ultralow-power, low-cost, and on-device inference. Their approach aims to perform real-time equipment monitoring and condition-based maintenance, addressing the limitations of traditional AI-based PdM systems that rely on resource-intensive servers.
The LOPdM IoT network was designed to operate in a star configuration, where multiple inference sensor nodes communicated with a central receiver via Bluetooth Low Energy (BLE). These sensor nodes comprised SPS sensors, an Analog-to-Digital Converter (ADC), a MCU, and a BLE module for data transmission. The SPS was a piezoelectric cantilever used to capture vibration signals; and the MCU preprocessed the data using Fast Fourier Transform (FFT), performed anomaly detection, and sent the results through the BLE module.
Considering the distorted nature of the vibration signals and the resource constraints of the MCU, the authors conducted a comprehensive comparison of classic ML models in a classification task. From this analysis, Random Forest and DL models were selected for deployment on the MCU. Due to the limited memory and processing power of the MCU, the DL model was compressed using the TFLM framework. After deployment, both models achieved high precision up to 99\%. Furthermore, the proposed SPS-based LOPdM system was compared against an Inertial Measurement Unit (IMU)- based approach commonly used for vibration monitoring. The results demonstrated that by leveraging SPS and TinyML, the LOPdM system achieved up to 67\% of energy savings compared to the traditional IMU-based system.

% PdM in Electrical Machines: An Edge Computing Approach
De las Morenas et al. \cite{related-work-sensor-inf-3} presented an edge computing based PdM framework for electrical machines. The framework focused on motor current signature analysis to detect broken rotor bars and eccentricity faults using FFT and a SVM classifier. This PdM solution was implemented on an Arduino platform, providing a cost-effective solution for small and medium sized enterprises. The proposed IoT network comprised sensor nodes, gateways, and cloud servers, but the primary computation was conducted at the edge, close to the physical machines. In this case, the sensor nodes were Arduino Uno boards equipped with non-invasive clamp sensors to measure the electric current consumed by the induction motors.
The system adopted a hybrid topology, combining elements of star and mesh networks to ease communication and data transfer between sensors and processing units. The communication within the framework employed UART-LAN/WIFI interfaces to manage data transmission without overloading the Arduino MCU. The preprocessing of data and feature extraction were performed on the Arduino, using dedicated Arduino libraries for FFT computation. The framework’s evaluation was conducted on a lathe turning machine equipped with a three-phase induction motor. The system successfully detected eccentricity and broken rotor bar faults by analyzing the sideband frequencies around the main power supply frequency (50 Hz). The specific frequencies indicative of faults were 44.667 Hz for broken rotor bars and 26.333 Hz, 73.667 Hz, 121 Hz, and 168.333 Hz for eccentricity faults.
Performance metrics for the PdM framework included the accuracy of the SVM classifier for healthy and faulty motor states. The SVM classifier was trained using predefined healthy and faulty feature values, ensuring robust fault detection. The system demonstrated satisfactory results, confirming the health status of the tested motor by the absence of fault-indicative frequency peaks.

\subsubsection{Challenges and Gaps in Existing Work}

Regardless of the inference approach, certain challenges remain consistent across all related work. A main challenge is the rigidity of inference location within PdM frameworks. Traditional systems employ a fixed inference location, which can be detrimental in real-world scenarios where operational conditions fluctuate rapidly. This lack of flexibility highlights the need for adaptive condition monitoring systems that can switch between inference strategies on-the-run, either by manual intervention or automatically through intelligent decision-making mechanisms. Secondly, the existing literature often overlooks critical network management aspects such as device provisioning and node orchestration.
Addressing these gaps, the proposed ESN-PdM framework aims to combine the strengths of cloud-based, gateway-based, and sensor-based inference approaches, establishing a hierarchical structure that can dynamically adjust the inference location based on user input or an adaptive inference heuristic. Additionally, the framework incorporates network management functionalities to ensure device provisioning, node orchestration, and efficient data communication. Table \ref{table:related-work-comparison}  offers a comparative analysis of existing approaches versus the ESN-PdM framework proposed in this paper.

\begin{table*}[ht]
\centering
\caption{\textbf{Comparison of existing approaches versus the ESN-PdM framework.}}
\label{table:related-work-comparison}
\renewcommand{\arraystretch}{2.5} % Adjust the vertical padding here
\resizebox{\textwidth}{!}{
    \begin{tabular}{||>{\centering\arraybackslash}m{1.8cm}||
        >{\centering\arraybackslash}m{2cm}|
        >{\centering\arraybackslash}m{2cm}|
        >{\centering\arraybackslash}m{2cm}||
        >{\centering\arraybackslash}m{1.5cm}|
        >{\centering\arraybackslash}m{1.5cm}|
        >{\centering\arraybackslash}m{1.5cm}||
        >{\centering\arraybackslash}m{1.8cm}|
        >{\centering\arraybackslash}m{1.8cm}||}
        \hline
        \multirow{3}{*}{} & \multicolumn{3}{c||}{\textbf{PdM Framework Features}} & \multicolumn{3}{c||}{\textbf{Inference Location}} & \multicolumn{2}{c||}{\textbf{PdM Model}} \\ \cline{2-9} 
        & \textbf{Node Provisioning} & \textbf{Real-time Data Collection} & \textbf{Node Orchestration} & \textbf{Cloud} & \textbf{Gateway} & \textbf{Sensor} & \textbf{ML} & \textbf{DL} \\ [0.5ex]
        \hline \hline
        {\cite{related-work-cloud-inf-1}} & & \checkmark & & \checkmark & & & & \checkmark \\ 
        {\cite{related-work-cloud-inf-2}} & & \checkmark & & \checkmark & & & \checkmark & \\ 
        {\cite{related-work-cloud-inf-3}} & & \checkmark & \checkmark & \checkmark & & & & \checkmark \\ 
        {\cite{related-work-gateway-inf-1}} & & \checkmark & & & \checkmark & & & \checkmark \\ 
        {\cite{related-work-gateway-inf-2}} & \checkmark & \checkmark & & & \checkmark & & & \checkmark \\ 
        {\cite{related-work-gateway-inf-3}} & & \checkmark & \checkmark & & \checkmark & & \checkmark & \\ 
        {\cite{related-work-sensor-inf-1}} & & \checkmark & & & & \checkmark & & \checkmark \\ 
        {\cite{related-work-sensor-inf-2}} & & \checkmark & & & & \checkmark & & \checkmark \\ 
        {\cite{related-work-sensor-inf-3}} & & & \checkmark & & & \checkmark & \checkmark & \\ \hline
        \textbf{Our Framework} & \checkmark & \checkmark & \checkmark & \checkmark & \checkmark & \checkmark & & \checkmark \\ \hline
    \end{tabular}
    }
\end{table*}

\section{ESN-PdM Framework Overview}
\label{sec:esn-pdm-framework-overview}

The ESN-PdM framework is based on a microservice architecture, a design pattern in which loosely coupled services are organized around business capabilities. This architecture style promotes the development of complex applications as a suite of small, independent services, each running in its own process. This design allows the framework to be highly scalable, flexible, and modular, making it suitable for various industrial applications. The framework is organized into three distinct layers, each corresponding to a specific device type: the Cloud layer, Gateway layer, and Sensor layer. Figure 2 provides an overview of the IoT Condition Monitoring System, illustrating the interactions between the different devices in the framework. The sensor nodes collect real-time data and send it wirelessly to the gateways, which manage the nodes and route the data to a Cloud Service Provider (CSP) for storage and analysis. Each device in the framework is equipped with a specific ML or DL model tailored to its computational capabilities, enabling efficient condition monitoring and anomaly detection.

\Figure[ht](topskip=0pt, botskip=0pt, midskip=0pt)[width=1.95\columnwidth]{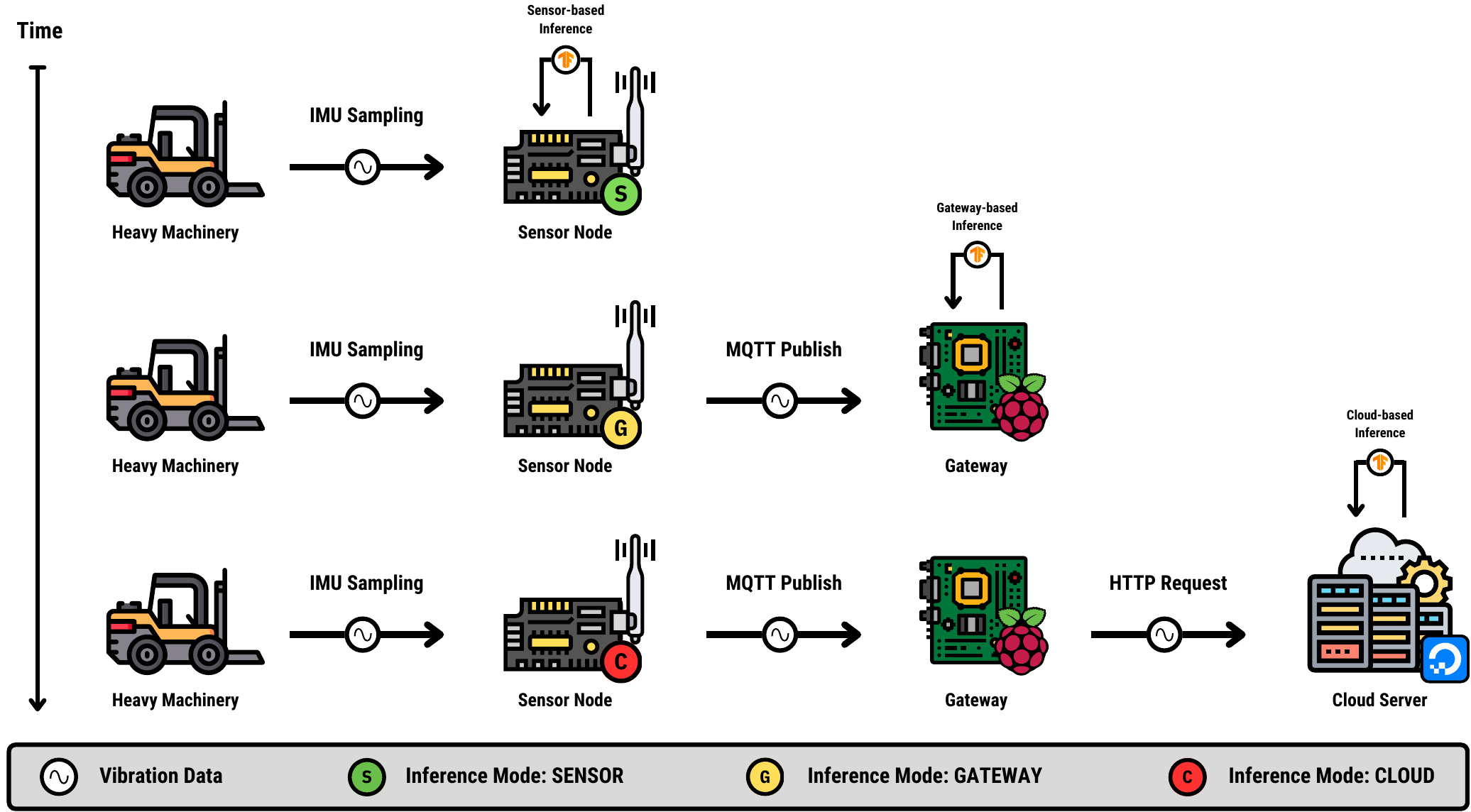}{ \textbf{Overview of the components of proposed ESN-PdM framework. The system consisted in three inference layers: sensor nodes, gateway, and a cloud server.}\label{fig:iot-condition-monitoring}}

\subsection{Cloud Layer}
The Cloud layer operates on cloud infrastructure such as Digital Ocean, AWS, Azure, or Google Cloud. It manages incoming requests from applications and gateways, routing them to the appropriate microservices. Its key functions include network management (discovering, provisioning, and configuring sensors), data storage and retrieval, and command execution. Figure \ref{fig:cloud-layer} shows the Cloud layer architecture, consisting of the Cloud API, Data Microservice, Command Microservice, and Inference Microservice.

\Figure[ht](topskip=0pt, botskip=0pt, midskip=0pt)[width=0.95\columnwidth]{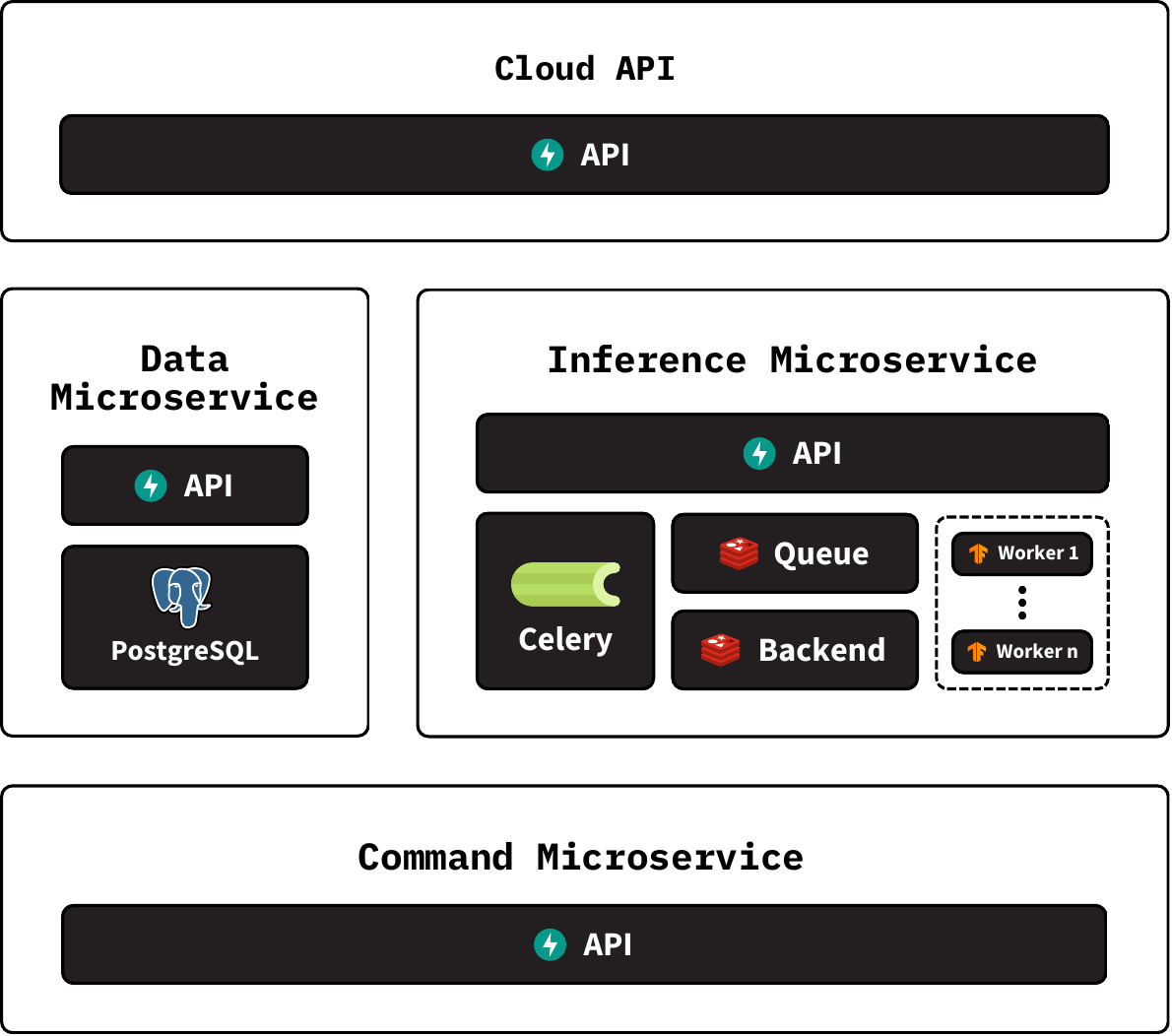}{ \textbf{The cloud layer of the ESN-PdM framework. It is composed of the Cloud API, Data Microservice, Command Microservice, and Inference Microservice.}\label{fig:cloud-layer}}

\subsubsection{Cloud API}
The Cloud API serves as the entry point for all external and internal requests to the Cloud layer. It provides a RESTful interface for interacting with the rest of the cloud services, enabling users to manage sensors and gateways, retrieve data, and execute commands. The API is built using FastAPI, a lightweight python web framework that leverages the Starlette ASGI framework/toolkit for asynchronous request handling. Additionally, the API is secured using JWT tokens for authentication and authorization, ensuring that only authorized users can access the system’s resources.

\subsubsection{Data Microservice}
The Data Microservice serves as the entry point for all data-related operations in the Cloud layer. It is responsible for storing both structured metadata key for the network’s operation such as gateway and sensor IDs, property values and state, as well as unstructured data such as sensor readings, parameters and inference results. The microservice is built on top of a PostgreSQL database, a powerful open-source object-relational database system known for its reliability, robustness, and performance. The Data Microservice exposes a set of RESTful endpoints for CRUD operations on the database, enabling the Cloud API to interact with the underlying data store.

\subsubsection{Command Microservice}
The Command Microservice is responsible for managing the network’s devices by sending commands to gateways and sensors. It provides an interface for operating on device properties such as setting the sleep period for sensors, updating the TensorFlow (TF) model for inference, and retrieving the list of provisioned nodes. A device property is a well-defined attribute of a device that can be read or written to, it has a specific data type, allows certain methods, and is associated with a target device type. Table \ref{table:device-properties} lists the main device properties available in the Command Microservice. 

\begin{table*}[ht]
    \centering
    \caption{Device Properties table available in the Command Microservice of proposed Cloud Layer.}
    \label{table:device-properties}
    \renewcommand{\arraystretch}{1.5} % Adjust the vertical padding here
    \resizebox{\textwidth}{!}{
        \begin{tabular}{>{\centering\arraybackslash}m{3.5cm}
            >{\centering\arraybackslash}m{5.5cm}
            >{\centering\arraybackslash}m{2.5cm}
            >{\centering\arraybackslash}m{1.5cm}
            >{\centering\arraybackslash}m{1.5cm}
            >{\centering\arraybackslash}m{1.5cm}}
        \toprule\toprule
        \textbf{Property Name} & \textbf{Content} & \textbf{Allowed Methods} & \textbf{Target Device}  & \textbf{Data Type}\\ \hline
        
        \texttt{tf\_model\_bytes} & TF model gzipped and encoded in base64 & SET & Gateway / Sensor & \texttt{uint8*} \\ \hline
        \texttt{tf\_model\_size} & TF model size in bytes & SET & Gateway / Sensor & \texttt{uint32} \\ \hline
        \texttt{provisioned\_nodes} & List of provisioned nodes & SET / GET / ADD & Gateway & \texttt{char**} \\ \hline
        \texttt{gateway\_id} & Gateway ID & GET & Gateway & \texttt{char*} \\ \hline            
        \texttt{sensor\_id} & Sensor ID & GET & Sensor & \texttt{char*} \\ \hline
        \texttt{sleep\_period} & Sleep time for each sensor cycle & SET / GET & Sensor & \texttt{uint32} \\ \hline
        \texttt{state} & Sensor State & SET / GET & Sensor & \texttt{uint32} \\ \hline
        \texttt{inference\_mode} & Sensor Inference Mode & SET / GET & Sensor & \texttt{uint8} \\ \toprule\toprule
        \end{tabular} 
    }
\end{table*}

\subsubsection{Inference Microservice}
The Inference Microservice is responsible for executing ML/DL models on the cloud server. It consists of three main components: the API, an asynchronous task queue, and a cluster of inference workers. The API serves as an interface for uploading TF models, submitting inference requests, and retrieving results. The asynchronous task queue, managed by Celery, schedules and distributes inference tasks to the available workers. The inference workers then execute these tasks using the uploaded models and store the results in a Redis database for efficient retrieval.  Figure \ref{fig:inference-microservice} illustrates the data flow within the Inference Microservice, highlighting the interactions between the API, task queue, and workers.

\Figure[ht](topskip=0pt, botskip=0pt, midskip=0pt)[width=0.95\columnwidth]{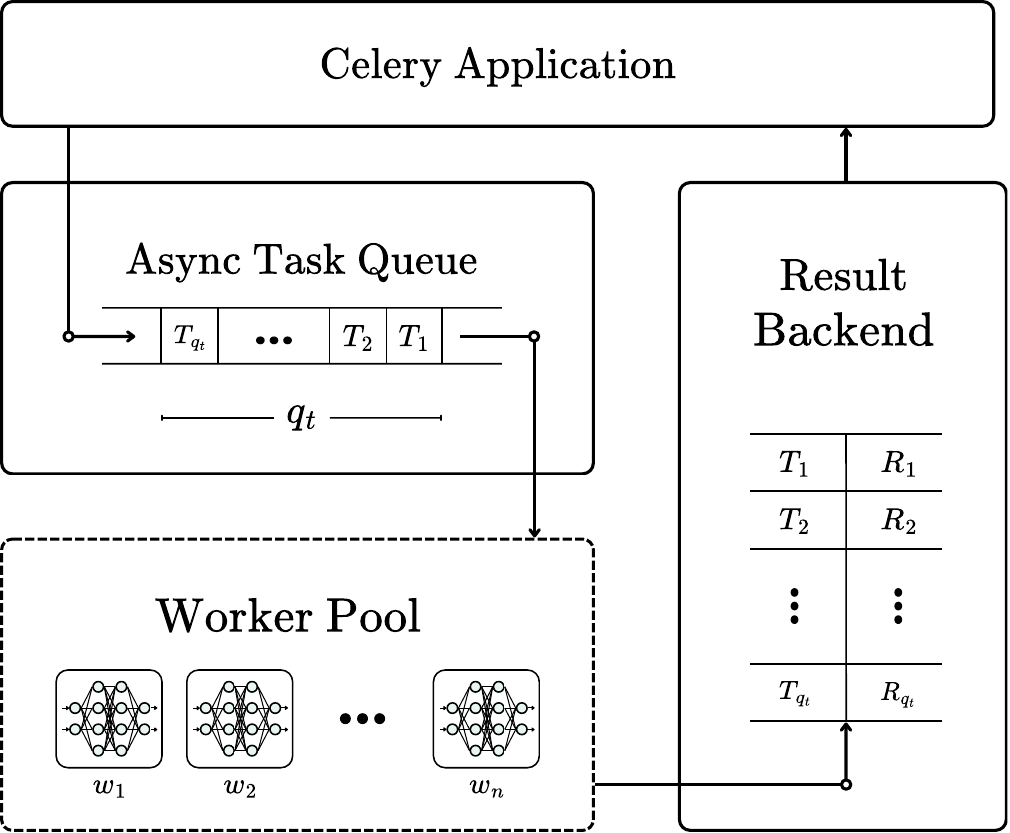}{ \textbf{Diagram of the proposed Inference Microservice. It depicts the dataflow through the API, the asynchronous task queue, and the inference worker.}\label{fig:inference-microservice}}

\subsection{Gateway Layer}
The Gateway layer bridges the Cloud layer and the Sensor layer. Each gateway functions as the central hub for a WSN built on a WiFi network it hosts. The gateway is implemented using a Raspberry Pi 4 Model B with 8GB of RAM, a powerful SBC featuring a quad-core 64-bit ARM Cortex-A72 processor running at 1.5 GHz. The device includes integrated dual-band 2.4 GHz and 5 GHz IEEE 802.11ac Wi-Fi and Bluetooth 5.0 BLE connectivity, both using a shared antenna for efficient wireless communication. Ethernet connectivity is also available for wired communication, enabling the gateway to access the internet and communicate with the cloud server. Although the Raspberry Pi 4 offers low power consumption, the device was powered directly from the mains in this study to ensure continuous operation.

Key responsibilities of the Gateway layer include discovering and setting up sensors, handling read/write operations, relaying data between sensors and the cloud, and orchestrating the network's nodes. Figure \ref{fig:gateway-layer} illustrates the Gateway layer architecture, comprising the Gateway API, Metadata Microservice, Inference Microservice, MQTT-Sensor Microservice and BLE Provisioning Microservice.

\Figure[ht](topskip=0pt, botskip=0pt, midskip=0pt)[width=0.95\columnwidth]{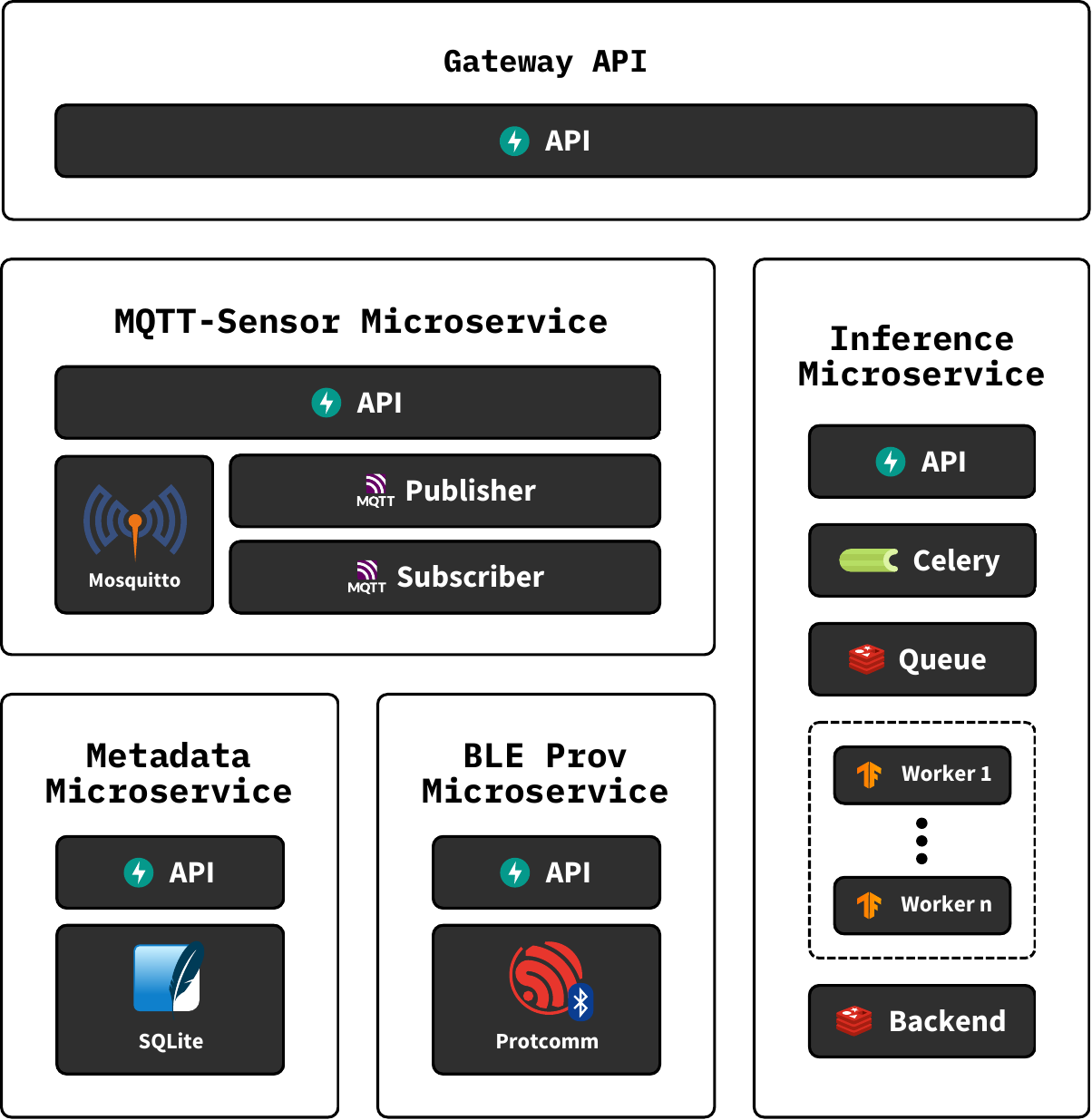}{ \textbf{Gateway layer of the proposed ESN-PdM framework. It is composed by the Gateway API, Metadata Microservice, MQTT-Sensor Microservice, Inference Microservice, and BLE Provisioning Microservice. }\label{fig:gateway-layer}}

Among the microservices in the Gateway layer, the Gateway API, the Metadata Microservice, and the Inference Microservice share the same structure and functionalities as their counterparts in the Cloud layer. The Gateway API handles all external and internal requests directed to the Gateway layer. The Metadata Microservice manages the metadata of the network’s devices, and the Inference Microservice executes ML/DL models on the gateway. The remaining microservices, the MQTT-Sensor Microservice and the BLE Provisioning Microservice, are unique to the Gateway layer and are described in detail below.

\subsubsection{MQTT-Sensor Microservice}
The MQTT-Sensor Microservice is responsible for managing the communication between the gateway and the sensor nodes through the MQTT protocol. Besides the MQTT Broker (Eclipse Mosquitto), the microservice consists of two main components: the MQTT Publisher and the MQTT Subscriber. The MQTT Publisher exposes a RESTful API for the Gateway API to send commands to the sensor nodes through MQTT. This is achieved through the FastAPI-MQTT library, which integrates MQTT functionality into the FastAPI framework. On the other hand, the MQTT Subscriber listens for incoming messages from sensor nodes and forwards them to the Gateway API for processing. This include sensor readings, device status updates and command responses.

\subsubsection{BLE Provisioning Microservice}
The BLE Provisioning Microservice handles the discovery and provisioning of new sensor nodes over BLE. This process involves configuring the sensor nodes with the WiFi credentials required to join the gateway’s WSN. The provisioning process is based on Espressif Protocomm provisioning protocol, a secure and efficient method for configuring IoT devices over BLE. It begins with the microservice creating a Protocomm instance and registering security and data handlers that manage secure connections and handle incoming provisioning commands. In this case, the provisioning device is the gateway that establishes a connection to the node over BLE. Then, a secure session is established between the two devices through a two-way handshake. Once a secure channel is established, the gateway sends the WiFi credentials to the sensor node, which then joins the WSN and terminates the BLE connection. Figure \ref{fig:ble-provisioning} illustrates the BLE provisioning process, showing the four main stages: device discovery, session establishment, configuration, and connection termination.

\Figure[ht](topskip=0pt, botskip=0pt, midskip=0pt)[width=0.95\columnwidth]{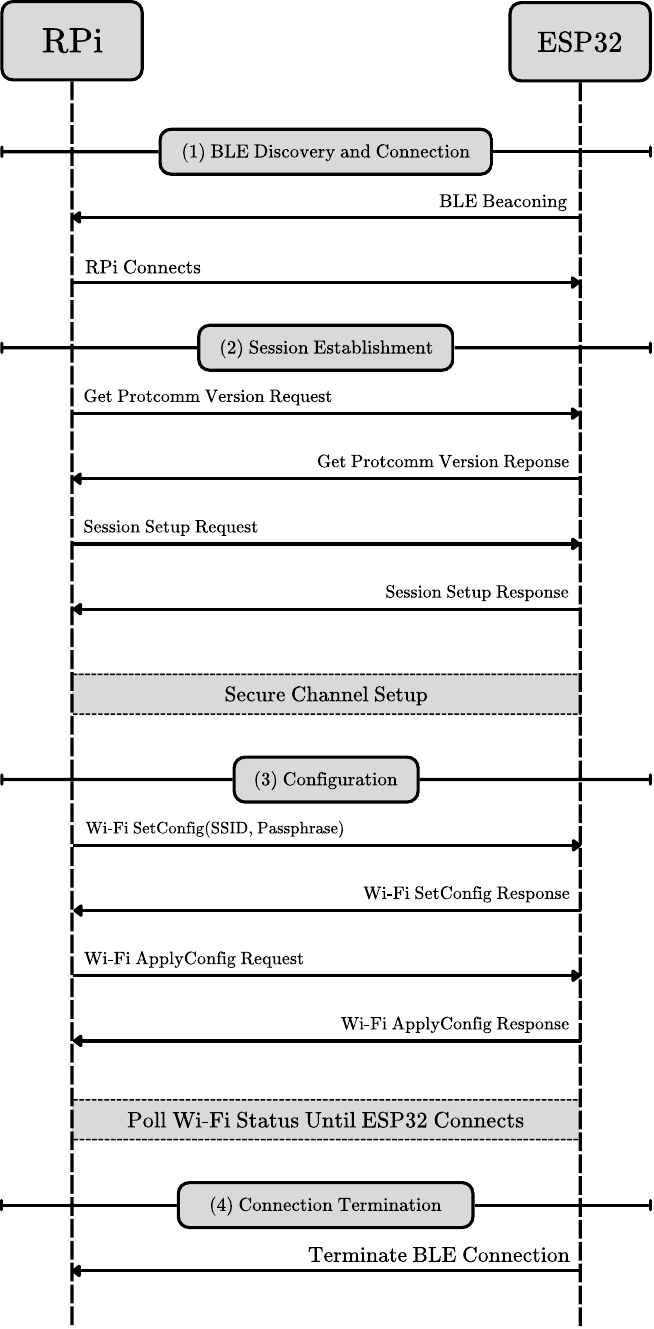}{ \textbf{Espressif's protcomm provisioning protocol over BLE. It is composed of four main stages: (1) device discovery, (2) session establishment, (3) configuration, and (4) connection termination.}\label{fig:ble-provisioning}}

Espressif Protocomm provisioning protocol supports three security schemes, each providing a different level of security for the provisioning process. For the ESN-PdM framework, Security Scheme 1 was chosen due to its balance between security and performance (see Figure~\ref{fig:security-scheme1}). During the session starting phase, both the gateway and sensor node generate their own Elliptic Curve Diffie-Hellman (ECDH) key pairs using Curve25519. They then exchange public keys, and each side computes a shared secret using its private key and the other party’s public key. If authentication through a Proof-of-Possession (PoP) was selected, the shared secret is combined with a hash of the PoP value to derive a session key. The session key is then used with AES256-CTR to encrypt and authenticate the exchanged public keys, ensuring that both parties have knowledge of the shared secret to establishing a secure communication channel for the subsequent provisioning process.

\Figure[ht](topskip=0pt, botskip=0pt, midskip=0pt)[width=0.95\columnwidth]{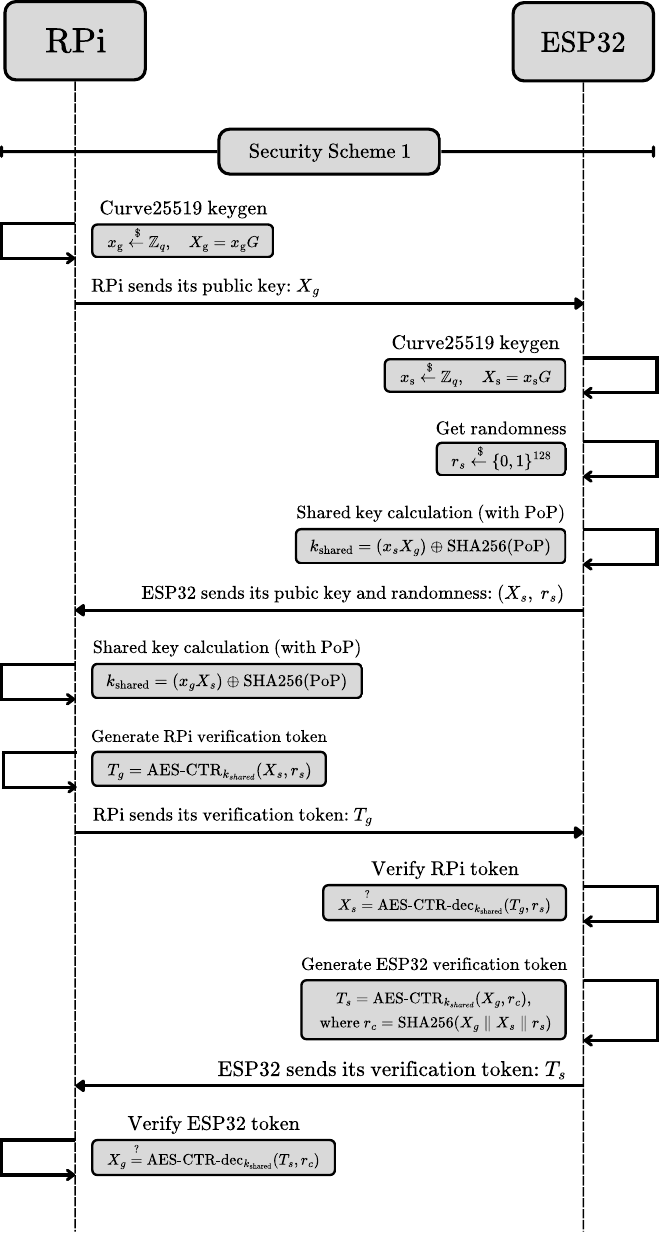}{ \textbf{Security scheme 1 for the protcomm provisioning protocol, leverages Curve25519 for key exchange, AES256-CTR for encryption/decryption, and a PoP for authentication.}\label{fig:security-scheme1}}

\subsection{Sensor Layer}
The Sensor layer forms the foundation of the ESN-PdM framework, directly interfacing with the physical environment. This layer consists of sensor nodes deployed across the industrial environment, to collecting data and transmitting it wirelessly to the upper layers if required. Figure \ref{fig:sensor-node-hardware} shows the hardware design of the sensor node, developed around the ESP32-WROOM-32 MCU, a dual-core processor based on the Tensilica Xtensa LX6 architecture operating at up to 240 MHz. This MCU integrates Wi-Fi and Bluetooth (v4.2 BR/EDR and BLE) connectivity, making it ideal for IoT applications that require wireless data transmission. With 520 KB of SRAM and up to 4 MB of external flash memory, it provides sufficient resources for processing and storing sensor data. The ESP32 supports multiple communication interfaces such as UART, SPI, and I²C, facilitating seamless integration with external sensors like the Bosch BMI270 IMU. The BMI270 is a 6-axis IMU comprising a 3-axis accelerometer and a 3-axis gyroscope, capable of measuring accelerations from ±2g to ±16g and angular rates from ±125°/s to ±2000°/s. Communicating via the I²C interface, the BMI270 offers a data output rate of up to 6.4 kHz, suitable for capturing high- frequency vibrations, and includes onboard motion detection algorithms to reduce processing overhead on the MCU. A microSD card slot is also integrated into the sensor node for data logging, enabling the storage of sensor readings in case of network disruptions. The sensor node is powered by a 1400 mAh lithium-polymer (LiPo) battery, known for its high energy density and compact form factor, ensuring long operating times and portability.

\Figure[ht](topskip=0pt, botskip=0pt, midskip=0pt)[width=0.95\columnwidth]{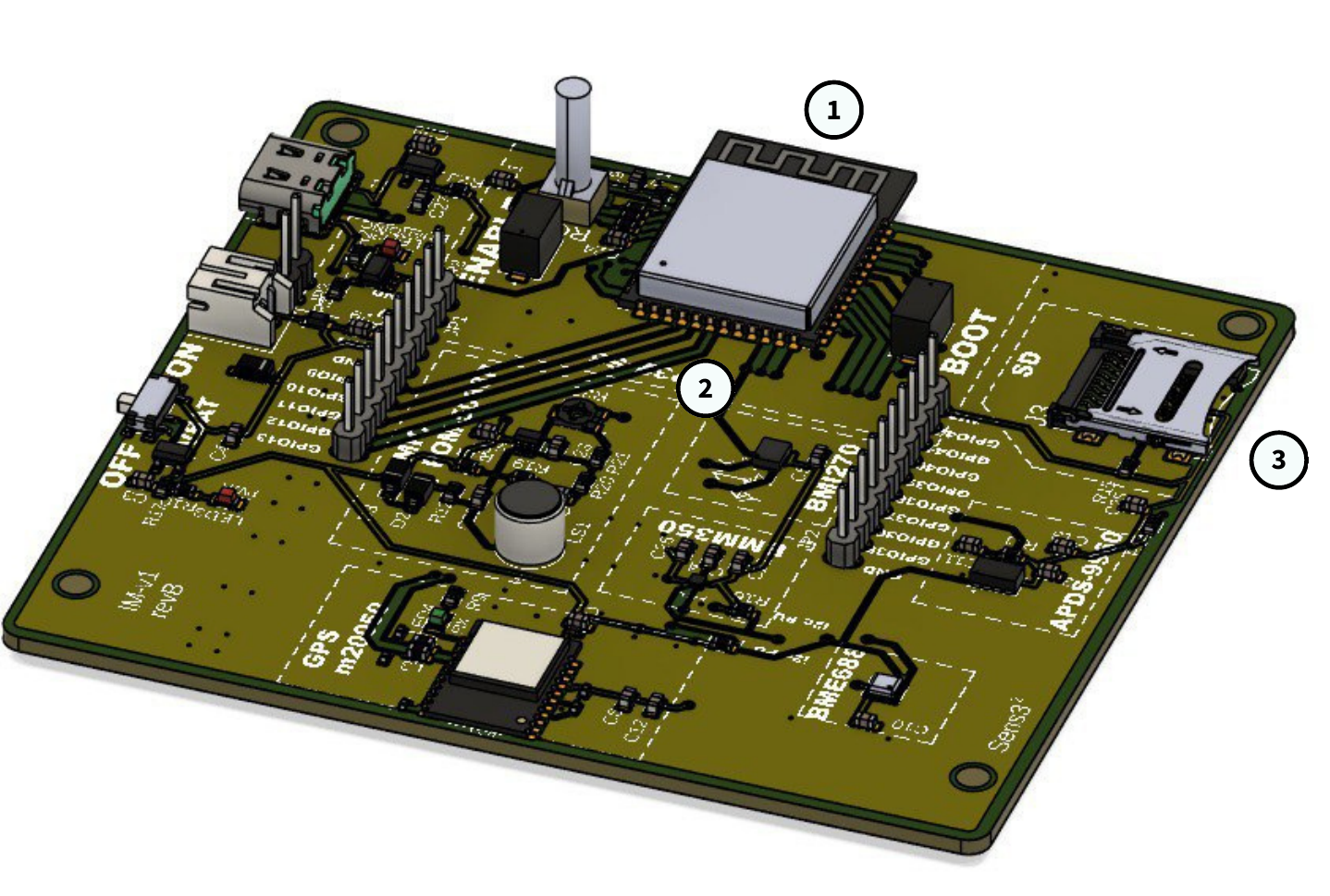}{ \textbf{3D Render of the sensor node hardware, showcasing the ESP32 MCU (1), the Bosch BMI270 IMU (2) and the microSD card slot (3)}\label{fig:sensor-node-hardware}}

Built on top of the Espressif IoT Development Framework (ESP-IDF), the firmware of the sensor node provides a robust platform for data collection and on-device inference. It is optimized for low power consumption, efficient data trans- mission, and seamless integration with upper layers, while maintaining a flexible, modular architecture. As depicted in \ref{fig:sensor-node-firmware}, the firmware comprises several components: (i) BLE provisioning, (ii) MQTT communication, (iii) TFLM inference, (iv) data collection, (v) compression /decompression, and (vi) sensor state management.

The BLE Provisioning component handles the device side of the Protocomm provisioning protocol, enabling the ESP32 to securely obtain Wi-Fi credentials. The MQTT Communication component allows the sensor node to publish and subscribe to MQTT topics, ensuring seamless data exchange with the gateway layer. This component depends on the Compression/Decompression component, which uses a version of Zlib optimized for embedded systems, to compress and decompress all packets exchanged between the sensor and gateway layers. The Inference component uses the TFLM framework and ESP-NN, an open-source library for Espressif chips containing optimized implementations of kernels used in TFLM. The Data Collection component interfaces with the BMI270 IMU, employing the I²C serial protocol to configure registers and read the sensor data.

\Figure[ht](topskip=0pt, botskip=0pt, midskip=0pt)[width=0.95\columnwidth]{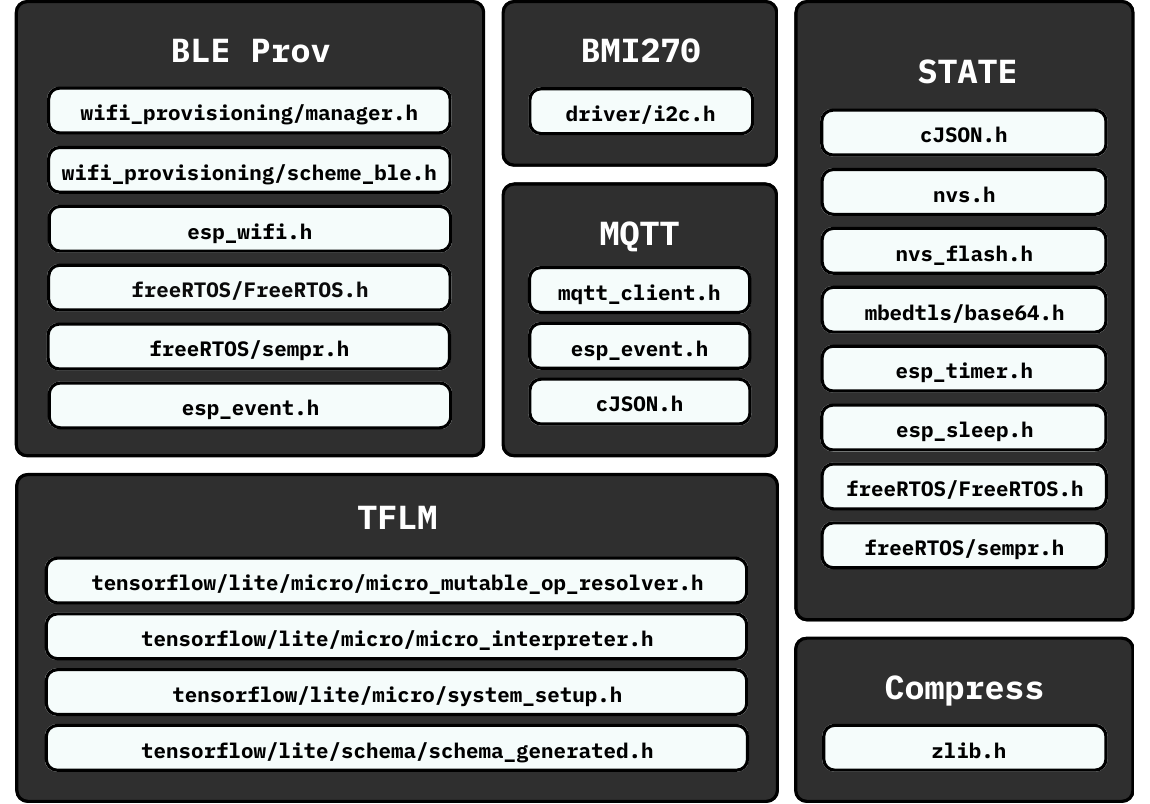}{ \textbf{Firmware diagram of the sensor node, composed of components for BLE provisioning, MQTT communication, Inference, data collection, compression/decompression, and sensor state management.}\label{fig:sensor-node-firmware}}

The State Management component is at the top of firmware design. It manages the sensor node’s core properties and behavior through two structures: the sensor node struct, which contains all data required for operation—such as the TFLM model, BMI270 configuration, Wi-Fi credentials, MQTT topics, and any writable/readable properties—and the node state machine. As shown in Figure 10, the state machine defines five possible states for the sensor node: INITIAL, UNLOCKED, LOCKED, WORKING, and IDLE. Transitions between these states are triggered by SET commands received from the gateway layer.
The node operates on a duty cycle consisting of a sleep phase and an active phase. Before entering the sleep phase, it serializes the sensor node structure and saves it to NVS for persistence across reboots. On the other hand, the active phase varies depending on the current state. For instance, during the INITIAL state, the node beacons for BLE provisioning. In the UNLOCKED state, it waits for the gateway to update its properties. In the LOCKED state, it listens for a configuration confirmation message to transition to the WORKING state, or a configuration rejection message to revert to the UNLOCKED state. During the IDLE state, the node remains inactive, awaiting a reset command from the gateway layer.
The most critical state is the WORKING state, where the node spends most of its time. The node’s behavior changes based on the inference mode. If the inference mode is set to GATEWAY or CLOUD, the node collects data from the BMI270 IMU, compresses it, turns on the radio, connects to the MQTT broker and publishes it. If the inference mode is set to SENSOR, it only collects data and runs the TFLM model locally. Since input from the gateway is vital for network orchestration, the node periodically turns on the radio when sensor is setup in the inference mode to process any pending commands. This approach allows the radio to remain off during most on-device inference cycles, turning on only when necessary, thereby conserving energy and extending the node’s battery life. 

\subsection{Adaptive Inference}
% Adaptive Inference Explaination
Adaptive Inference, a key feature of the ESN-PdM frame- work, allows sensor nodes to dynamically update their inference location based on operational conditions. This feature is enabled by two main components: the Inference Mode Property and the Adaptive Inference Heuristics. The Inference Mode Property specifies the target device to which the node sends its raw data for inference: S (Sensor), G (Gateway), or C (Cloud). The Adaptive Inference Heuristics analyze historical anomaly data and other factors to determine the optimal inference mode for a given sensor node. After each prediction, the target device runs an adaptive heuristic to determine the optimal inference mode for each node based on current information. If the recommended inference mode differs from the current mode, the device instructs the node to transition to the new mode.

The \textit{Sensor Heuristic} Heuristic decides whether to remain in  $S$ mode or switch to $G$ mode. The \textit{Gateway Heuristic} determines whether to stay in $G$ mode, switch back to $S$ mode, or escalate to $C$ mode. The \textit{Cloud Heuristic} decides whether to continue in $C$ mode, revert to $G$ mode, or force the node to de-escalate to $S$ mode. Figure~\ref{fig:inference-transitions} illustrates the possible transitions the inference mode can take based on the adaptive inference heuristics. 

\Figure[ht](topskip=0pt, botskip=0pt, midskip=0pt)[width=0.95\columnwidth]{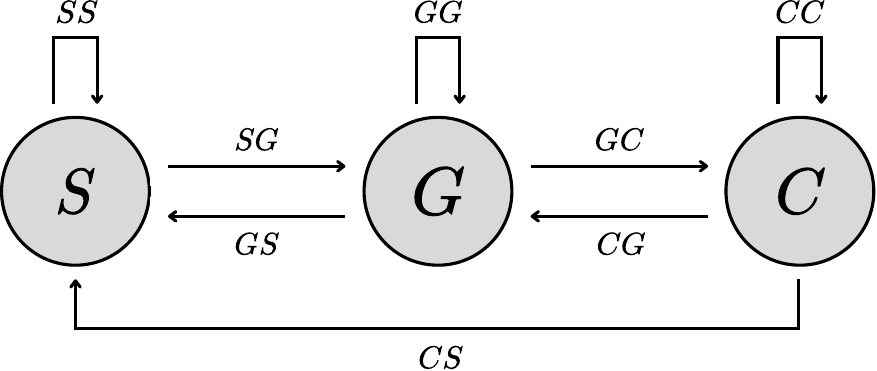}{ \textbf{Inference mode transitions in the ESN-PdM framework. The diagram illustrates possible transitions among sensor, gateway, and cloud modes based on adaptive heuristics.}\label{fig:inference-transitions}}

Nodes operate in a single inference mode at any given moment, thus only one heuristic is invoked at each time step. The heuristics are designed to optimize the trade-off between latency and prediction accuracy, escalating to higher layers when anomalies are frequently detected as accurate predictions are crucial, and reverting to lower layers when anomalies are infrequent to minimize latency.

A key factor in selecting the inference mode is the historical frequency of detected anomalies. Let $X_t$ denote the inference mode at time step $t$, where $X_t \in \{S, G, C\}$. For each node, every device type maintains an anomaly history, $H_t$, represented as a bitmask indicating whether an anomaly was detected at each time step from $t-h+1$ to $t$, where $h$ is the history memory depth. Equation \ref{eq:anomaly-history} shows how $H_t$ is updated based on the current prediction $p_t$, the previous history $H_{t-1}$, and the current mode $X_t$. Equation \ref{eq:history-length} defines the anomaly history length, $\tau_t$. Finally, Equation \ref{eq:anomaly-count} computes the anomaly count, $\sigma_t$, representing the number of anomalies detected in $H_t$. Figure \ref{fig:anomaly-history} provides a visual representation of the anomaly history computation process.

\Figure[ht](topskip=0pt, botskip=0pt, midskip=0pt)[width=0.80\columnwidth]{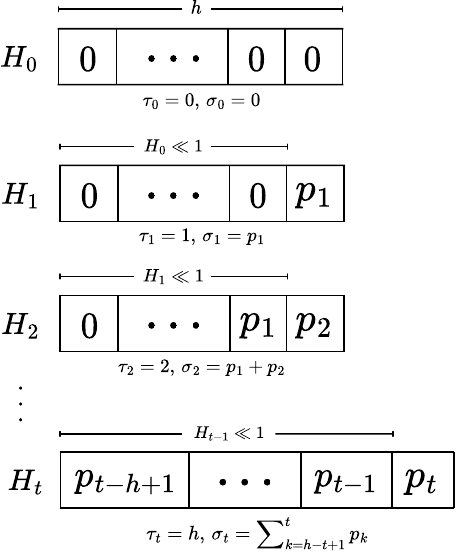}{ \textbf{Anomaly history computation. The diagram shows the process of updating the anomaly history, history length, and anomaly count for a given sensor node over time.}\label{fig:anomaly-history}}

\begin{equation}
    \label{eq:anomaly-history}
    % If X_t == X_t-1, then H_t = H_t-1 << 1 else 0
    H_t = \begin{cases}
        (H_{t-1} \ll 1) \ \& \ p_t & \text{if } X_t = X_{t-1} \\
        0 & \text{otherwise}
    \end{cases}
\end{equation}

\begin{equation}
    \label{eq:history-length}
    % If X_t != X_t-1, then tau_t = 0 else tau_t = tau(H_t-1) + 1
    \tau_t = \begin{cases}
        min(h, \tau_{t-1} + 1) & \text{if } X_t = X_{t-1} \\
        0 & \text{otherwise}
    \end{cases}
\end{equation}

\begin{equation}
    \label{eq:anomaly-count}
    % sigma_t = popcount(H_t)
    \sigma_t = \sum_{i=0}^{h-1} \left( \left( H_t \gg i \right) \,\&\, 1 \right)
\end{equation}

The adaptive inference heuristics utilize anomaly histories to optimize the selection of inference modes. When anomalies are infrequent, the node operates in lower layers to prioritize real-time processing and reduce latency. Conversely, if anomalies occur frequently, the system escalates to higher layers to enhance prediction accuracy and minimize false positives. The following sections detail the adaptive inference heuristics for each device type.

\subsubsection{Sensor Adaptive Heuristic}
Let $b_t$ denote the remaining battery level of the sensor node at time step $t$, where $b_t \in [0, 100]$, and $\psi_b$ represent the low battery threshold. Let $\psi_s$ be the abnormal prediction escalation threshold, indicating the maximum number of anomalies that can be detected before escalating to the gateway. Algorithm \ref{alg:sensor-adaptive-heuristic} outlines the Sensor Adaptive Inference Heuristic.

\begin{algorithm}[H]
    \caption{Sensor Adaptive Inference Heuristic}
    \label{alg:sensor-adaptive-heuristic}
    \begin{algorithmic}[1]
        \Require $(X_{t-1} = X_t = S) \land p_t, H_{t-1}, b_t, \psi_b, \psi_s, h$
        \Ensure The next inference mode $X_{t+1}$
        \State $H_t \gets (H_{t-1} \ll 1) \,\&\, p_t$
        \State $\tau_t \gets \min(h, \tau_{t-1} + 1)$
        \State $\sigma_t \gets \sum_{i=0}^{h-1} \left( \left( H_t \gg i \right) \,\&\, 1 \right)$

        \If{$b_t < \psi_b$} \Comment{Low battery, unable to escalate}
            \State \Return $S$
        \ElsIf{$\tau_t < h$} \Comment{Wait for history to fill}
            \State \Return $S$
        \ElsIf{$\sigma_t \geq \psi_s$}
            \State \Return $G$
        \Else
            \State \Return $S$
        \EndIf
    \end{algorithmic}
\end{algorithm}

\subsubsection{Gateway Adaptive Heuristic}
Let $\psi_g$ and $\phi_g$ be the abnormal prediction escalation and de-escalation thresholds, respectively, for the gateway. Let $q_t$ the size of the gateway's inference queue at time step $t$ and $psi_q$ the queue size threshold. Note each prediction request the gateway receives has attached the battery level $b_t$ of the sensor node sending the request. Algorithm \ref{alg:gateway-adaptive-heuristic} outlines the Gateway Adaptive Inference Heuristic.

\begin{algorithm}[H]
    \caption{Gateway Adaptive Inference Heuristic}
    \label{alg:gateway-adaptive-heuristic}
    \begin{algorithmic}[1]
        \Require $(X_{t-1} = X_t = G) \land p_t, H_{t-1}, q_t, \psi_g, \phi_g, \psi_q, h$
        \Ensure The next inference mode $X_{t+1}$
        \State $H_t \gets (H_{t-1} \ll 1) \,\&\, p_t$
        \State $\tau_t \gets \min(h, \tau_{t-1} + 1)$
        \State $\sigma_t \gets \sum_{i=0}^{h-1} \left( \left( H_t \gg i \right) \,\&\, 1 \right)$

        \If{$b_t < \psi_b$} \Comment{Low battery, immediately de-escalate}
            \State \Return $S$
        \ElsIf{$\tau_t < h$} \Comment{Wait for history to fill}
            \State \Return $G$
        \ElsIf{$\sigma_t < \phi_g$}
            \State \Return $S$
        \ElsIf{$\phi_g \leq \sigma_t < \psi_g \land q_t < \psi_q$}
            \State \Return $G$
        \Else
            \State \Return $C$
        \EndIf
    \end{algorithmic}
\end{algorithm}

\subsubsection{Cloud Adaptive Heuristic}
Let $\psi_c$ and $\phi_c$ be the abnormal prediction escalation and de-escalation thresholds, respectively, for the cloud. Similarly to the gateway heuristic, the cloud heuristic considers the power level $b_t$ of the sensor node sending the prediction request. Algorithm \ref{alg:cloud-adaptive-heuristic} outlines the Cloud Adaptive Inference Heuristic.

\begin{algorithm}[H]
    \caption{Cloud Adaptive Inference Heuristic}
    \label{alg:cloud-adaptive-heuristic}
    \begin{algorithmic}[1]
        \Require $(X_{t-1} = X_t = C) \land p_t, H_{t-1}, h$
        \Ensure The next inference mode $X_{t+1}$
        \State $H_t \gets (H_{t-1} \ll 1) \,\&\, p_t$
        \State $\tau_t \gets \min(h, \tau_{t-1} + 1)$
        \State $\sigma_t \gets \sum_{i=0}^{h-1} \left( \left( H_t \gg i \right) \,\&\, 1 \right)$

        \If {$b_t < \psi_b$} \Comment{Low battery, immediately de-escalate}
            \State \Return $S$
        \ElsIf{$\tau_t < h$} \Comment{Wait for history to fill}
            \State \Return $C$
        \ElsIf{$\sigma_t < \phi_c$}
            \State \Return $G$
        \Else
            \State \Return $C$
        \EndIf
    \end{algorithmic}
\end{algorithm}

\section{PdM CASE-STUDY}
\label{sec:case-study-mining-equipment}

Forklifts are critical machinery on mining operations, providing material handling capabilities and logistic support across expansive and rugged terrains. Their ability to transport heavy loads—including equipment, raw materials, and supplies—ensures smooth and uninterrupted processes. Given the harsh operating conditions, forklifts experience significant wear and tear over time, leading to potential malfunctions and breakdowns. It is well-documented that the vibration patterns of machinery can provide valuable insights into their operational health. By analyzing such data, it is possible to detect anomalies, predict failures, and optimize maintenance schedules. This scenario presents an ideal use case for the ESN-PdM framework, as it requires monitoring the operational condition of machinery in real-time based on vibration data. 

This section explores the fundamentals of DL and its application to the case-study proposed by providing a detailed description of the data acquisition process and data engineering, the DL model architecture, and the TinyML strategy adopted to deploying the models on hardware resource-constrained devices such as gateways and sensor ndoes. 

\subsection{DL Fundamentals}

DL, a subfield of ML, has gained popularity due to its ability to automatically learn complex patterns from raw data, eliminating the need for manual feature engineering. The foundation of DL lies in ANNs, computational models inspired by the structure and function of the human brain. ANNs consist of interconnected layers of units called neurons. Each neuron computes an activation $y$, as shown in Equation \ref{eq:neuron-output}, by applying weights $w_i$ to its inputs $x_i$, adding a bias $b$, and passing the result through a non-linear activation function $f$. For a neuron to learn, learnable parameters such as weights and biases need to be iteratively adjusted to minimize a loss function, a process known as training.

\begin{equation}
    \label{eq:neuron-output}
    y = f\left(\sum_{i=1}^{n} w_i x_i + b\right)
\end{equation}

Figure~\ref{fig:ann-diagram} illustrates a simple ANN with an input layer $\vec{x}$, multiple hidden layers $\vec{h}^{(i)}$, and an output layer $\vec{y}$. ANNs are typically organized in a feedforward manner, where the output of one layer serves as the input to the next. This motivates the vectorized representation of the ANN output in Equations \ref{eq:vectorized-ann-output}, where $L$ is the number of hidden layers, $\vec{x}$ is the input vector, $\vec{h}^{(i)}$ is the output of layer $i$, $W^{(i)}$ is the weight matrix for layer $i$, $b^{(i)}$ is the bias vector for layer $i$, and $f$ is the activation function.

\Figure[ht](topskip=0pt, botskip=0pt, midskip=0pt)[width=0.95\columnwidth]{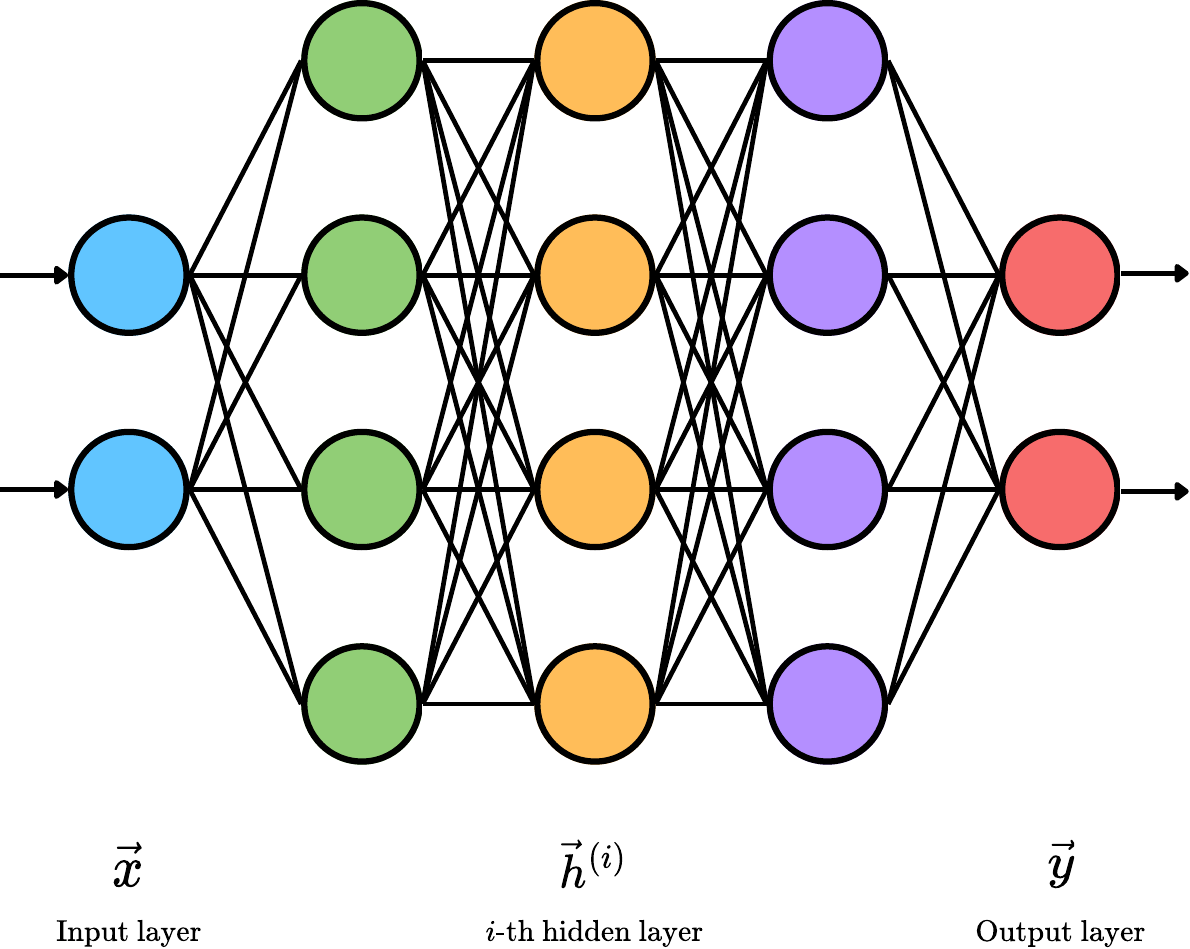}{ \textbf{Diagram of a simple ANN. It consists of an input layer, multiple hidden layers, and an output layer.}\label{fig:ann-diagram}}

\begin{equation}
    \label{eq:vectorized-ann-output}
    \begin{split}
        \vec{h}^{(0)} &= f(W^{(0)} \vec{x} + \vec{b}^{(0)}) \\
        \vec{h}^{(i)} &= f(W^{(i)} \vec{h}^{(i-1)} + \vec{b}^{(i)}) \\
        \vec{y} &= softmax(W^{(L)} \vec{h}^{(L-1)} + \vec{b^{(L)}})
    \end{split}
\end{equation}

Although ANNs have been around for decades, their popularity surged in the early 2010s due to advancements in computational power, the availability of large datasets, and the development of new architectures such as CNNs and RNNs.

CNNs are particularly effective for processing digital signals like audio, image and video. Their core operation, convolution, uses filters to extract patterns from input data, producing feature maps that highlight detected features. Non-linear activation functions are then applied to these feature maps to introduce non-linearity, and pooling layers reduce spatial dimensions, enabling the extraction of higher-level features. Figure~\ref{fig:cnn-diagram} illustrates a simple CNN that employs 1D convolutions for processing vibration signals commonly found in condition monitoring applications.

\Figure[ht](topskip=0pt, botskip=0pt, midskip=0pt)[width=0.95\columnwidth]{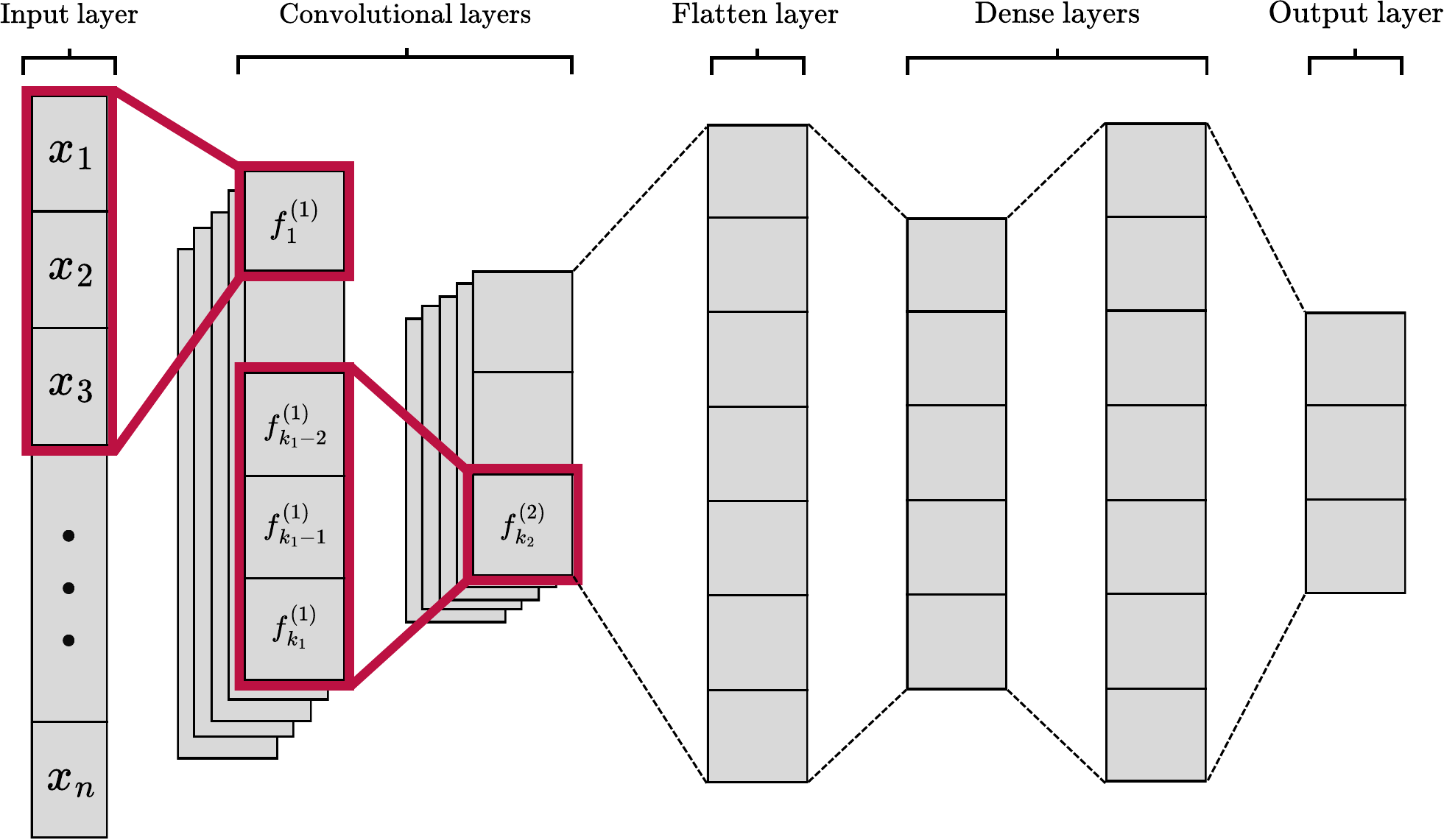}{ \textbf{Diagram of a simple CNN. It consists of an input layer, convolutional layers, a flattening layer, and a dense output layer.}\label{fig:cnn-diagram}}

RNNs, on the other hand, are designed to handle sequential data, such as time-series signals and natural language. They leverage recurrent connections to maintain a memory of past inputs, enabling them to capture temporal dependencies in the data. LSTMs are a special kind of RNN capable of learning long-term dependencies through the use of gating mechanisms that regulate the flow of information. These gates decide which information to keep, write, or erase at each time step, allowing the network to maintain and update a cell state that carries information across many time steps. Figure~\ref{fig:lstm-diagram} illustrates an LSTM cell with its input gate $i_t$, forget gate $f_t$, output gate $o_t$, cell state $c_t$, cell state update $\tilde{c}_t$, and hidden state $h_t$. Equation~\ref{eq:lstm-equations} provides the mathematical formulation of an LSTM cell, detailing the computations involved in updating the cell state $c_t$ and hidden state $h_t$ at time step $t$.

\Figure[ht](topskip=0pt, botskip=0pt, midskip=0pt)[width=0.95\columnwidth]{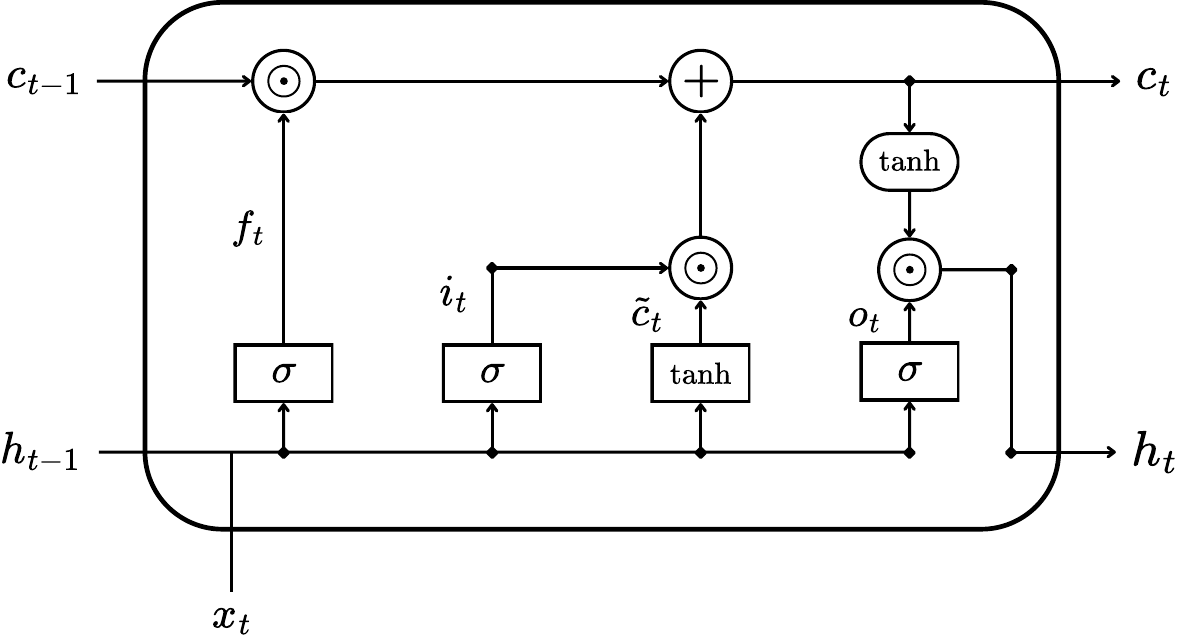}{ \textbf{Diagram of an LSTM cell. The cell consists of input, output, forget, and cell state gates, enabling it to learn long-term dependencies in sequential data.}\label{fig:lstm-diagram}}

\begin{equation}
    \label{eq:lstm-equations}
    \begin{split}
    f_t &= \sigma(W_f \cdot [h_{t-1}, x_t] + b_f) \\
    i_t &= \sigma(W_i \cdot [h_{t-1}, x_t] + b_i) \\
    \tilde{c}_t &= \tanh(W_c \cdot [h_{t-1}, x_t] + b_c) \\
    o_t &= \sigma(W_o \cdot [h_{t-1}, x_t] + b_o) \\
    c_t &= f_t \odot c_{t-1} + i_t \odot \tilde{c}_t \\
    h_t &= o_t \odot \tanh(c_t)
    \end{split}
\end{equation}

\subsection{Data Acquisition and Dataset Construction}
The data acquisition process began by defining normal operation for the target equipment, a Toyota 8FG45N forklift. According to OSHA standards, normal operation for a forklift involves safe use by trained and certified operators using properly maintained equipment within its rated capacity and in appropriate environments, following all prescribed safety procedures. Under these conditions, the forklift’s vibration patterns were measured using a reference accelerometer, the PCE-VDL 16I, which has a measurement range of ±18G and a frequency range of 10 Hz to 10 kHz. 
In the testing scenario, the vibration frequency during normal operation was found to be 25 Hz, with peak-to-peak amplitude variations of ±1.5G. Based on these preliminary measurements, the sensor node’s IMU was configured to sample at a rate of 50 Hz to capture the forklift’s vibration patterns effectively. The BMI270 IMU captured both 3-axis acceleration and 3-axis angular velocity data over 10-second sampling windows, constituting six time-series signals of 500 samples each per minute. At the end of each sampling window, the sensor node stored the data on a microSD card for further labeling after the data collection process.
The forklift had accumulated over 9,000 operational hours at the time of the study, with an expected service life of 10,000 to 15,000 hours depending on the maintenance sched- ule. After a period of three months, the acquired time-series signals were labeled based on diagnostic reports provided by the maintenance engineer on the day of signal acquisition, rather than event labeling. The labels were classified into four categories: Good (0), Acceptable (1), Unsatisfactory (2), and Unacceptable (3), representing the forklift’s operational health status. The final dataset consisted of over 25,000 labeled sequences, with 60\% used for training, 20\% for validation, and 20\% for testing. Table \ref{table:dataset-structure} summarizes the structure of the dataset.

% Table: Dataset Structure
% Column name | Description | C-like Type
\begin{table}[ht]
    \centering
    \caption{\textbf{Structure of the forklift condition monitoring dataset. It consists of time-series signals from the forklift's vibration data, each labeled with an operational health label.}}
    \label{table:dataset-structure}
    \renewcommand{\arraystretch}{1.5} % Adjust the vertical padding here
    \begin{tabular}{c c c}
    \toprule\toprule
    \textbf{Column Name} & \textbf{Description} & \textbf{Type} \\ \hline
    \texttt{seq\_id} & Unique identifier for each sequence & \texttt{uint32} \\ \hline
    \texttt{seq\_time} & Timestamp of the sequence start & \texttt{datetime64} \\ \hline
    \texttt{sample\_id} & Unique identifier for each sample & \texttt{uint16} \\ \hline
    \texttt{acc\_x} & Acceleration along the x-axis & \texttt{float32} \\ \hline
    \texttt{acc\_y} & Acceleration along the y-axis & \texttt{float32} \\ \hline
    \texttt{acc\_z} & Acceleration along the z-axis & \texttt{float32} \\ \hline
    \texttt{gyr\_x} & Angular velocity along the x-axis & \texttt{float32} \\ \hline
    \texttt{gyr\_y} & Angular velocity along the y-axis & \texttt{float32} \\ \hline
    \texttt{gyr\_z} & Angular velocity along the z-axis & \texttt{float32} \\ \hline
    \texttt{label} & Operational health label & \texttt{uint8} \\ \toprule\toprule
    \end{tabular}
\end{table}

\subsection{DL Model Architecture}
The DL model architecture selected for this study is a variation of the Long Short-Term Memory Fully Convolutional Network (LSTM-FCN) proposed by Karim et al. \cite{lstmfcn}. The LSTM-FCN architecture effectively combines the strengths of LSTM networks and CNNs to capture both temporal and spatial dependencies in time-series data. This capability is particularly advantageous for the case study, as the vibration signals from the forklift are sequential and multi-dimensional. Another benefit of the LSTM-FCN architecture is its simplicity, consisting of well-established computational blocks. This simplicity makes it compatible with lightweight DL runtimes like TF Lite, which only support a subset of regular operations, facilitating deployment on hardware resource-constrained devices. Moreover, the LSTM-FCN is a model that has demonstrated high accuracy in various real-world applications.

The model comprises two primary routes for feature extraction: the LSTM route and the CNN route. The LSTM route processes the input sequence using an LSTM layer to capture long-term dependencies, with a dropout layer incorporated to prevent overfitting. In parallel, the CNN route employs a series of one-dimensional convolutional layers, each followed by batch normalization and ReLU activation functions, to extract spatial features from the input sequence. The outputs from both routes are then concatenated and passed through a global average pooling layer to reduce the feature map dimensions. The proposed model differs from the original LSTM-FCN by incorporating three fully connected layers with dropout and batch normalization for classification. Figure~\ref{fig:dl-model-architecture} illustrates the proposed architecture of the DL model.

\Figure[ht](topskip=0pt, botskip=0pt, midskip=0pt)[width=1.95\columnwidth]{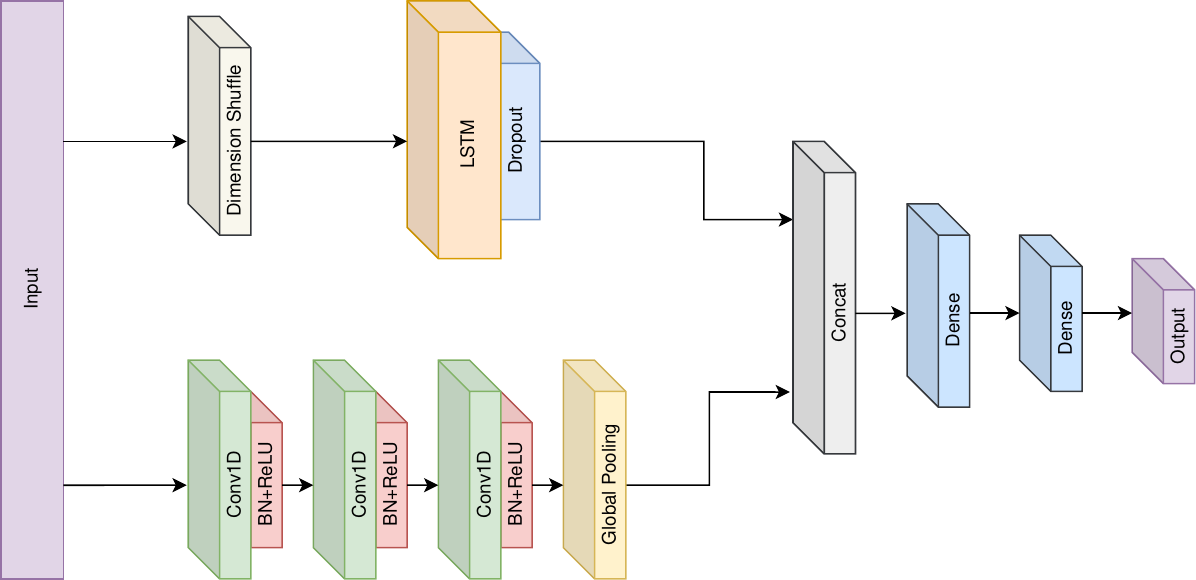}{ \textbf{Proposed architecture of the DL model for forklift condition monitoring. It consists of LSTM and CNN routes for feature extraction, followed by fully connected layers for classification.}\label{fig:dl-model-architecture}}

Since the proposed model must operate on all three layers of the ESN-PdM framework, three versions of the model were trained: a full sized model for the cloud (tagged as large), a medium sized model for the gateway (tagged as medium), and a small sized model for the sensor (tagged as tiny) Table~\ref{tab:model-arch-summary} summarizes the architecture of each model version, detailing the types of hidden layers and the number of units in each layer. Each model version was trained using the hyperparameters recommended by the authors of the LSTM-FCN model \cite{related-work-cloud-inf-1}.

\newcolumntype{C}[1]{>{\centering\arraybackslash}m{#1}}

\begin{table}[h]
\centering
\label{tab:model-arch-summary}
\caption{\textbf{Architecture summary of the DL models. Each version of the model is tailored to run on a specific layer of the ESN-PdM framework.}}
    \resizebox{\columnwidth}{!}{
    \begin{tabular}{
            C{2.0cm} % Model Version column
            C{2.0cm} % Type of Hidden Layer column
            C{3.0cm} % Number of Units column
        }
        \toprule\toprule
        \textbf{Model Version} & \textbf{Layer Type} & \textbf{Number of Units}  \\
        \midrule % Optional: Replace with \hline
        \multirow{3}{*}{Large} & LSTM         &  (64)               \\
                            & Conv1D        & (128, 256, 128)                 \\
                            & Dense       &   (128, 64)              \\
        \midrule % Optional: Replace with \hline
        \multirow{3}{*}{Medium} & LSTM       &  (32)                \\
                                & Conv1D        &  (64, 128, 64)                \\
                                & Dense        &  (64, 32)                \\
        \midrule % Optional: Replace with \hline
        \multirow{3}{*}{Tiny} & LSTM        &   (16)              \\
                            & Conv1D       & (32, 64, 32)                \\
                            & Dense        &   (32, 16)               \\
        \bottomrule\bottomrule
    \end{tabular}
    }
\end{table}

\subsection{TinyML Optimization}
To achieve optimal performance on hardware resource-constrained devices, merely defining specific model version sizes for each target device is insufficient. A comprehensive TinyML optimization strategy is key to reduce the model’s footprint, complexity, and computational demands to fit within the limitations of each device layer. The cloud layer operates on powerful servers with virtually no memory or computational constraints, allowing models to run directly on TF without optimization. In contrast, the gateway layer—implemented on a Raspberry Pi 4 Model B—has limited memory and lacks hardware acceleration capabilities, demanding the use of an optimized runtime like Tensor- Flow Lite for inference.
The sensor layer—implemented on an ESP32 MCU—presents the most stringent constraints, offering approximately 300 KB of available memory with only 120 KB being contiguous, a critical requirement since the TFLM runtime loads models into contiguous memory. Another factor to consider is that, regardless of the device layer, models are sent through the network in a compressed format. However, the effectiveness of the compression depends heavily on the presence of redundant parameters in the model’s architecture.

To address the memory footprint challenges of both the gateway and sensor models, the TF Lite converter was used to transform the DL models into smaller FlatBuffer files suitable for deployment. However, this conversion alone was insufficient for the stringent hardware constraints. Given that the Raspberry Pi 4 has significantly more memory than the ESP32, optimization strategies were tailored accordingly. For the gateway model, dynamic range quantization was applied, which quantizes only the weights to 8-bit integers while dynamically quantizing and dequantizing activations during inference. This method reduces the model size without requiring a representative dataset for calibration. 
For the sensor model, full integer quantization was employed, transforming all learnable parameters and neuron activations to 8-bit integers. This more aggressive quantization is necessary to fit the stringent memory limitations of the ESP32, albeit at the cost of reduced model accuracy and the need for calibration data.
To further increase compression effectiveness, post- training pruning was applied to both models. Pruning involves zeroing out a percentage of model parameters, introducing sparsity. The process requires defining initial and final sparsity levels and a schedule for increasing sparsity. A polynomial decay schedule was used. The initial sparsity was set at the default 25\%, and experiments were conducted to determine optimal final sparsity levels by evaluating model accuracy versus sparsity percentages. 

Figure~\ref{fig:pruning-accuracy} illustrates the trade-off between model accuracy and sparsity for both the gateway and sensor models. For the gateway model, a final sparsity of 65\% provides an acceptable performance while significantly reducing parameters. For the sensor model, a final sparsity of 50\% was selected for a similar balance. 
Although the pruned models remain the same size in bytes before compression, the increased number of zeroed parameters creates patterns that compression algorithms exploit more effectively. Consequently, the pruned models compress to smaller sizes, facilitating transmission and storage.

\Figure[ht](topskip=0pt, botskip=0pt, midskip=0pt)[width=0.95\columnwidth]{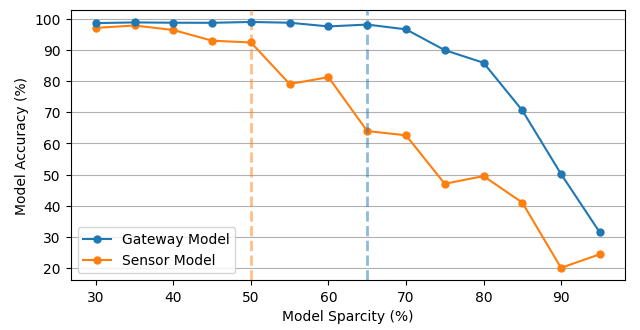}{\textbf{Model accuracy versus model sparsity for the gateway and sensor models. The graph highlights the optimal final sparsity levels for each model.}\label{fig:pruning-accuracy}}

Table~\ref{table:model-size-complexity} 5 compares the size and complexity of the cloud, gateway, and sensor models after applying the previously described TinyML optimization techniques. The gateway model, when gzipped, is approximately 50 times smaller than the cloud model, while the sensor model is approximately 170 times smaller than the cloud model and roughly three times smaller than the gateway model. These significant reductions in model size and complexity demonstrate the effectiveness of the TinyML optimization strategy in tailoring models to the constraints of each device layer.

\begin{table}[ht]
    \centering
    \caption{\textbf{Model size and complexity comparison after TinyML optimizations. The table shows the total number of parameters, the size of the model in kilobytes (KB), and the gzipped size of the model in KB.}}
    \label{table:model-size-complexity}
    \renewcommand{\arraystretch}{1.5} % Adjust the vertical padding here
    \resizebox{\columnwidth}{!}{
        \begin{threeparttable}
            \begin{tabular}{c c c c}
            \toprule\toprule
            \textbf{PdM Model} & \textbf{Total parameters} & \textbf{Size (KB)} & \textbf{Gzipped Size (KB)} \\ \hline
            Cloud Model & 252,868 & 3,032 & 2,755 \\ \hline
            Gateway Model & 64,996 & 90 & 56 \\ \hline
            Sensor Model & 17,140 & 30 & 16 \\ \toprule\toprule
            \end{tabular}
        \end{threeparttable}
    }
\end{table}

\section{Experimental Evaluation}
\label{sec:framework-evaluation}
A comprehensive experimental evaluation was conducted to assess the performance of the ESN-PdM framework in the context of forklift condition monitoring. The evaluation focuses on the classification performance of the proposed DL models, the energy consumption of the sensor node for both onboard and offboard inference scenarios, and the inference latency measured by the sensor node during real-time operation, triggering the shifting between inference modes based on the adaptive heuristics.

\subsection{Classification Performance}
The operational condition classification performance of the DL models was evaluated using the test subset of the forklift condition monitoring dataset detailed in the previous section. Models were assessed based on their accuracy and recall scores for each class: Good, Acceptable, Unsatisfactory, and Unacceptable. As expected, the Cloud Model outperformed the Gateway and Sensor Models, achieving the highest overall accuracy of 99.38\%. It maintained consistently high recall rates across all classes, particularly surpassing in the Good (98.11\%) and Unacceptable (99.69\%) categories. The Gateway Model, with an accuracy of 94.06\%, delivered robust recall scores, especially in the Unsatisfactory class with a 98.09\% recall rate. The Sensor Model achieved an accuracy of 91.40\%, demonstrating strong recall in the Acceptable (99.41\%) and Un acceptable (99.61\%) classes.
These performance differences can be attributed to the varying model sizes and optimization strategies employed for each target device. Overall, the results indicate that even with reduced parameters and optimized configurations, the Gateway and Sensor Models maintain high classification effectiveness, making them suitable for deployment in hardware resource-constrained environments. The evaluation results are summarized in 
Table~\ref{table:performance-models}.

\begin{table*}[ht]
    \centering
    \caption{\textbf{Performance of the condition monitoring models. The table shows the accuracy and recall scores for each class of the cloud, gateway, and sensor models.}}
    \label{table:performance-models}
    \renewcommand{\arraystretch}{1.5} % Adjust the vertical padding here
    \begin{tabular}{>{\centering\arraybackslash}m{2.5cm}
        >{\centering\arraybackslash}m{2.5cm}
        >{\centering\arraybackslash}m{2.5cm}
        >{\centering\arraybackslash}m{2.5cm}
        >{\centering\arraybackslash}m{2.5cm}
        >{\centering\arraybackslash}m{2.5cm}}
    \toprule\toprule
    \textbf{PdM Model} & \textbf{Accuracy (\%)} & \textbf{Recall Good class (\%)} & \textbf{Recall Acceptable class (\%)} & \textbf{Recall Unsatisfactory class (\%)} & \textbf{Recall Unacceptable class (\%)} \\ \hline

    Cloud Model  & 99.38   & 98.11  & 99.71  & 98.56 & 99.69  \\ \hline
    Gateway Model & 94.06   & 84.40  & 96.77  & 98.09 & 95.11  \\ \hline
    Sensor Model & 91.40 & 81.13   & 99.41  & 97.81  & 99.61  \\ \toprule\toprule
    \end{tabular}
\end{table*}    

\subsection{Energy Consumption}
Energy consumption is a critical factor in the ESN-PdM framework, particularly for sensor nodes that rely on battery power. Depending on the inference mode, a node’s behavior—and correspondingly its energy consumption—can varies significantly. In Sensor S mode, the node performs onboard inference, using energy solely for data sampling and processing. Conversely, in Gateway G and Cloud C modes, inference is offloaded to the respective layers, replacing TFLM-based inference with signal compression and subsequent radio transmission.

The experimental setup for measuring the energy consumption of the sensor node consisted on an ESP32-WROOM-32 Devkit, a SparkX Qwiic BMI270 breakout board, and Nordic Semiconductor’s Power Profiler Kit II (PPK2). The PPK2 is a power optimization tool for embedded systems that accurately measures and optionally supplies currents ranging from sub-microamperes up to 1 ampere, making it ideal for profiling power consumption in IoT devices. 

For experimental evaluation, the PPK2 was powered using a single USB cable and configured in Source Measure Unit (SMU) mode to supply up to 500 mA of current at 3.7 V to the ESP32-WROOM-32 Devkit, simulating a typical LiPo battery voltage. The PPK2 was connected to the VIN pin of the ESP32. The SparkX Qwiic BMI270 breakout board was powered by the ESP32’s 3.3 V pin and connected via the I2C bus for data communication. Additionally, the BMI270 IMU was configured to sample acceleration and angular velocity data at 50 Hz, with each sample window lasting approximately 10 seconds.

Figure~\ref{fig:power-consumption} illustrates the current consumption of the sensor node over time, detailing the power profiles for onboard inference cycles (Sensor S mode) and offboard inference cycles (Gateway G or Cloud C modes). In both scenarios, key operations were separated by 5 s delays to isolate their current consumption. Equation~\ref{eq:energy-consumption}was used to calculate the energy consumption, integrating the product of the current consumption and the supply voltage over the time interval of each operation.

% Equation for the energy consumption calculation
\begin{equation}
    \label{eq:energy-consumption}
    E = \int_{t_0}^{t_f} V(t)I(t) \, dt
\end{equation}

\Figure[ht](topskip=0pt, botskip=0pt, midskip=0pt)[width=0.95\columnwidth]{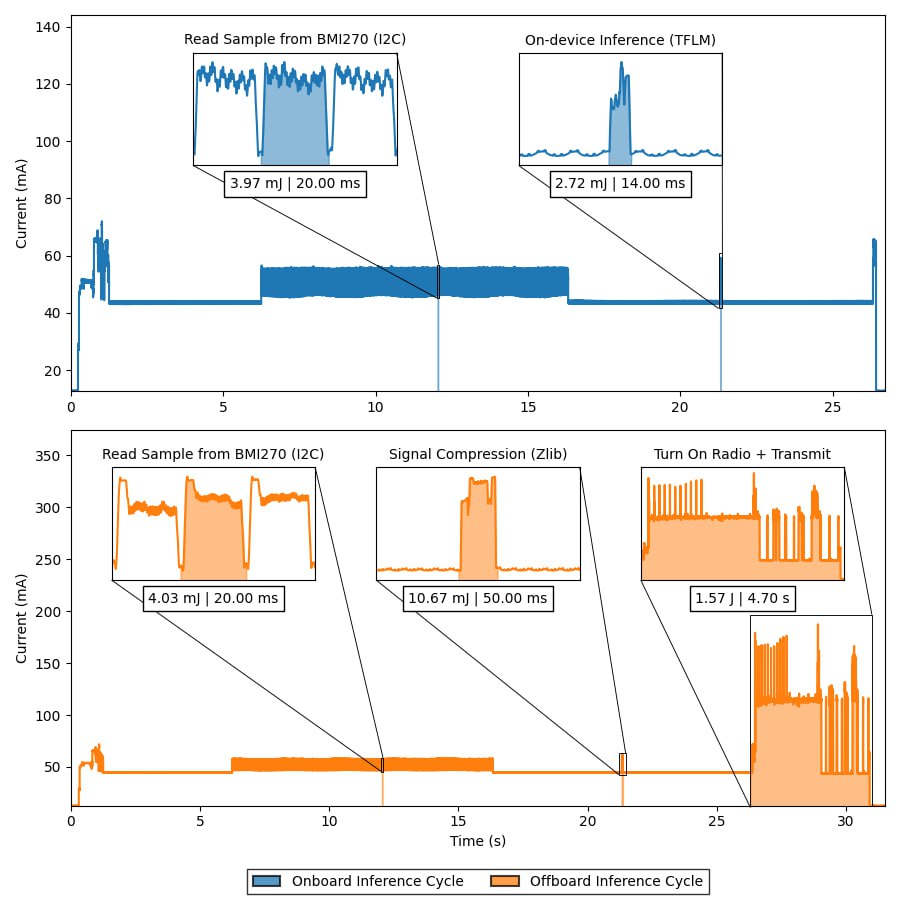}{
    \textbf{Energy consumption of the sensor node over time, detailing power profiles for onboard inference cycles in SENSOR mode and offboard inference cycles in GATEWAY or CLOUD modes.}
    \label{fig:power-consumption}
}

% Energy Consumption per Operation
The sensor node consumed approximately $4.00\,\text{mJ}$ of energy over $20\,\text{ms}$ to read a single 6-dimensional IMU sample (12 bytes). Collecting 500 samples resulted in a total energy consumption of $500 \times 4.00\,\text{mJ} = 2000\,\text{mJ}$, reading $5.86\,\text{KB}$ of data in about 10\,s. Data processing reveals a significant difference in energy consumption between onboard and off-board inference scenarios. For onboard inference, the sensor node quantizes raw data to 8-bit integers, producing a $2.92\,\text{KB}$ tensor, which is then fed to the TFLM model for inference. The entire process consumes $2.72\,\text{mJ}$ over $14\,\text{ms}$. In contrast, offboard inference in GATEWAY and CLOUD modes involves compressing the signal with Zlib, which, for a $5.86\,\text{KB}$ signal, consumes $10.67\,\text{mJ}$ over $50\,\text{ms}$. Additionally, radio transmission is required to send the compressed signal to the gateway or cloud, consuming $1570\,\text{mJ}$ over $4700\,\text{ms}$ for a compressed signal size of approximately $3\,\text{KB}$. Table~\ref{table:energy-consumption-summary} summarizes the energy consumption for each measured operation.

\begin{table}[ht]
    \centering
    \caption{\textbf{Summary of the measured operations. The table shows the data size, duration, and energy consumption for data sampling, TFLM inference, signal compression, and radio transmission.}}
    \label{table:energy-consumption-summary}
    \renewcommand{\arraystretch}{1.5} % Adjust the vertical padding here
    \resizebox{\columnwidth}{!}{
        \begin{threeparttable}
            \begin{tabular}{c c c c}
            \toprule\toprule
            \textbf{Measured Operation} & \textbf{Data Size (KB)} & \textbf{Duration (ms)} & \textbf{Energy (mJ)} \\ \midrule
            Data Sampling & 5.86 & 10000 & 2000.00 \\ \midrule
            TFLM Inference & 2.92 & 14 & 2.72 \\ \midrule
            Signal Compression & 5.86 & 50 & 10.67 \\ \midrule
            Radio Transmission & 3.00 & 4700 & 1570.00 \\ \bottomrule\bottomrule
            \end{tabular}
        \end{threeparttable}
    }
\end{table}

% Active and Sleep Phase Energy Consumption
Based on the previous results, the active phase of an onboard inference cycle consumes $2000.00\,\text{mJ} + 2.72\,\text{mJ} = 2002.72\,\text{mJ}$ over $10000\,\text{ms} + 14\,\text{ms} = 10014\,\text{ms}$. Conversely, the active phase of an offboard inference cycle consumes $2000.00\,\text{mJ} + 10.67\,\text{mJ} + 1570.00\,\text{mJ} = 35780.67\,\text{mJ}$ over $10000\,\text{ms} + 50\,\text{ms} + 4700\,\text{ms} = 14750\,\text{ms}$. For the sleep phase, the ESP32 consumes $10\,\mu\text{A}$ in deep sleep. Assuming a duration of 30\,s, the node consumes $10\,\mu\text{A} \times 30\,\text{s} \times 3.7\,\text{V} = 1.11\,\text{mJ}$. Thus, a full cycle in SENSOR mode consumes $2002.72\,\text{mJ} + 1.11\,\text{mJ} = 2003.83\,\text{mJ}$ over $10014\,\text{ms} + 30\,\text{s} = 40014\,\text{ms}$, while a full cycle in GATEWAY or CLOUD mode consumes $35780.67\,\text{mJ} + 1.11\,\text{mJ} = 35781.78\,\text{mJ}$ over $14750\,\text{ms} + 30\,\text{s} = 44750\,\text{ms}$.

% Battery Life Calculation for Both Scenarios
Since the sensor node switches between onboard and off- board inference modes using adaptive heuristics, determining an exact battery life is impractical. Instead, lower and upper bounds for battery life are established. The lower bound assumes only on-board inference, while the upper bound assumes only off-board inference. Equation~\ref{eq:battery-life} calculates battery life, where $E_{\text{battery}}$ is the total energy stored in the battery, $E_{\text{cycle}}$ is the energy consumed per cycle, and $t_{\text{cycle}}$ is the cycle duration.

\begin{equation}
    \label{eq:battery-life}
    \text{Battery Life} = \frac{E_{\text{battery}}}{E_{\text{cycle}}} \times t_{\text{cycle}} 
\end{equation}

Considering a standard LiPo battery with a capacity of 1,400 mAh and a voltage of 3.7 V, the total energy stored in the battery is $1{,}400\,\text{mAh} \times 3.7\,\text{V} = 5{,}180\,\text{mWh}$. Since $1\,\text{mWh} = 3.6\,\text{J}$, the total energy stored in the battery converts to $5{,}180\,\text{mWh} \times 3.6\,\text{J/mWh} = 18{,}648\,\text{J}$. Likewise, the total energy consumption per cycle was converted to joules, where $2{,}003.83\,\text{mJ} = 2.00383\,\text{J}$ and $3{,}581.78\,\text{mJ} = 3.58178\,\text{J}$. Therefore, the lower bound for the battery life is calculated as $18{,}648\,\text{J} / 3.58178\,\text{J} \times 44{,}750\,\text{ms} \approx 65\,\text{hours}$. Similarly, the upper bound is $18{,}648\,\text{J} / 2.00383\,\text{J} \times 40{,}014\,\text{ms} \approx 104\,\text{hours}$. These results demonstrate that the sensor node can operate for extended periods, even with the energy-intensive off-board inference mode.

The energy consumption in onboard inference mode was approximately 2003.83 mJ per cycle, while offboard inference mode consumption was 3578.67 mJ per cycle due to the additional energy cost of wireless data transmission. The percentage savings were calculated as follows:

\[
\text{Energy Savings (\%)} = \left( \frac{3578.67 - 2003.83}{3578.67} \right) \times 100\% \approx 44\%
\]

This approach shows that performing inference locally on the sensor node significantly reduces energy consumption, extending the node's operational life up to 104 hours in low-power mode. This is critical for PdM applications in remote industrial environments.

\subsection{Inference Latency}

% Inference Latency 
Inference latency is a critical performance metric in ML-enabled WSNs, representing the time elapsed between a sensor node sending a prediction request and receiving a corresponding response from an upper layer. Thus, inference latency directly impacts the responsiveness of the ESN-PdM framework, influencing the timeliness of maintenance actions and the overall system efficiency. Moreover, parameterizing the evolution of inference latency over time is essential for understanding the framework’s adaptability to changing operational conditions and the trade-offs between energy efficiency and response times.

% Experimental Setup
The experimental setup consisted of a sensor layer composed of one single sensor node deployed in a mining warehouse in Chile, simulating real-world operational conditions. The gateway layer was composed of Raspberry Pi 4 Model B installed in the control room at the warehouse, directly connected to internet via Ethernet. The cloud layer was implemented using DigitalOcean Droplets, hosted in a data center in NY, USA, providing scalable computational resources for processing prediction requests. To emulate realistic operator behavior, the sensor node was programmed to simulate forklift operators, exporting data with an anomaly probability of 0.3 based on the forklift condition monitoring dataset.

% Inference Latency Measurement
Since the ESN-PdM framework generates responses only emitted when the adaptive heuristic triggers an update in the inference mode, accurately measuring the evolution of inference latency over time is challenging as many prediction requests do not result in a response. To address this challenge, the experiment incorporated a mechanism where a blank response was sent to the sensor node after each prediction request that does not initiate an inference mode update. This modification enabled the sensor node to consistently measure the inference latency when dealing with upper layers by simply subtracting the inference request send time attached to the response from the current time.

% Adaptive Heuristic Parameters
The experiment considered ideal conditions concerning battery life and queue sizes for the gateway and cloud inference microservices, as adapting the inference mode based on anomaly detection was the primary focus. Foe each layer, the adaptive heuristic parameters were methodically configured. The anomaly history length was set to 32 for the sensor, 16 for the gateway, and 8 for the cloud. The anomaly thresholds were set to $\psi_s = 4$ for the sensor, $\phi_g = 4$ and $\psi_g = 8$ for the gateway, and $\phi_c = 2$ for the cloud. Since decreased history lengths and higher $\psi$ values correspond to lower tolerance for anomalies, the history lengths decrease and the $\psi$ values increase progressively up the hierarchy.

% Inference Latency Results
Figure~\ref{fig:inference-latency} illustrates the evolution of inference latency perceived by the sensor node over time. The experiment was conducted over a period of approximately 30 minutes, with the sensor node requiring a prediction every 10 seconds. The graph shows the latency for each inference mode: SENSOR, GATEWAY, and CLOUD. The SENSOR mode exhibits the lowest latency, with an average of 3.33\,ms, reflecting the direct inference performed on the sensor node. The GATEWAY mode presents higher latency, averaging 148.15\,ms, due to the overhead introduced by signal compression and radio transmission. However, it remains significantly lower than the CLOUD mode as communication is limited to the local network. The CLOUD mode exhibits the highest latency, averaging 641.71\,ms, reflecting the additional latency introduced by internet communication to an external server. Additionally, the graph highlights the adaptability of the ESN-PdM framework, with the sensor node dynamically shifting between inference modes based on the adaptive heuristics.

\Figure[ht](topskip=0pt, botskip=0pt, midskip=0pt)[width=0.95\columnwidth]{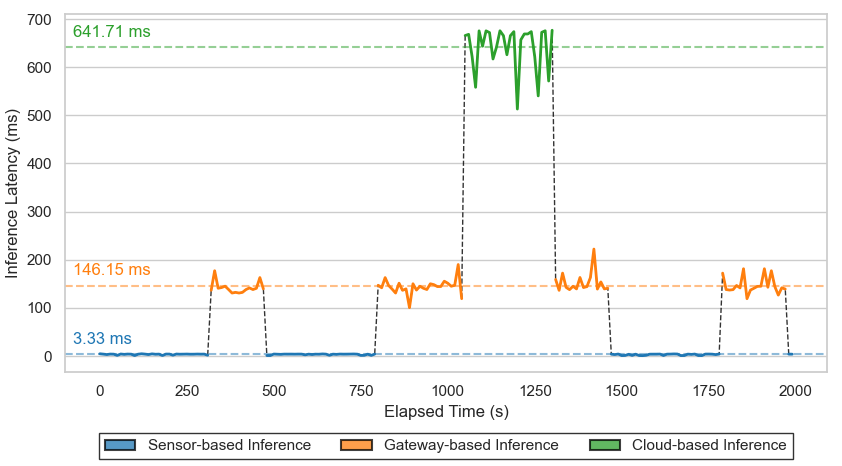}{ \textbf{Evolution of sensor node inference latency over time, illustrating the latency impact of switching between inference modes.}\label{fig:inference-latency}}

\section{Conclusion}
\label{sec:conclusion}

The ESN-PdM framework introduced in this study demonstrates a significant advancement in the PdM of industrial equipment operating in non-stationary environments, particularly within the mining sector. Through a hierarchical inference mechanism, the framework effectively balances latency, accuracy, and energy efficiency by dynamically selecting inference locations based on operational conditions. As a result, the ESN-PdM achieved high classification accuracy across all models, with the cloud model exhibiting the highest accuracy at 99.38\%, followed closely by the gateway and sensor models with accuracies of 94.06\% and 91.40\%, respectively. This classification performance indicates that, even in hardware-constrained environments, the framework maintains strong operational state recognition, making it suitable for real-time industrial applications.
An important aspect of the ESN-PdM framework is its capability for generating alarms at each layer—sensor, gateway, and cloud—based on real-time inference results. By generating alarms at the edge (sensor), the framework provides an immediate response to detected anomalies, allowing local operators to quickly address minor issues that may not require higher-level intervention. This low-latency response capability, with the sensor layer achieving an inference latency of 3.33 ms, is essential for real-time applications and supports immediate safety interventions when necessary. However, by offloading the alarm generation to the gateway or cloud under certain operational conditions, the system can leverage more sophisticated models with broader context and increased accuracy. For example, gateway-based inference can provide a balance between latency and computational power, offering alarms that integrate data from multiple sensors for a more comprehensive view, with a moderate latency of 148.15 ms.
The cloud-based inference, despite having the highest latency of 641.71 ms, offers the most advanced and accurate analysis, as it can aggregate extensive data from multiple sources. This layer is particularly suited for generating high-priority alarms that require in-depth analysis or predictive insights. Such alarms can indicate imminent failures, thus enabling PdM systems to provide advanced warnings well in advance of critical failures, supporting optimized maintenance scheduling and reducing downtime. The alarm generation hierarchy also allows the framework to selectively filter alarm types according to urgency and context, reducing false positives that often burden maintenance teams in traditional systems.
Energy efficiency is another crucial outcome of this study, especially for the sensor nodes deployed in energy-limited scenarios. The adaptive inference capabilities facilitated notable energy savings, as observed in the power consumption analysis. The sensor node's battery life was shown to range between approximately 65 hours in energy-intensive modes and 104 hours under more efficient conditions Such flexibility in power usage extends the operational time of sensor nodes, which is vital for continuous and autonomous PdM in remote and challenging locations.
Overall, the ESN-PdM framework contributes to the field of PdM by offering a versatile and scalable solution that addresses the inherent challenges of latency, accuracy, and energy efficiency in non-stationary industrial environments. By enabling adaptive alarm generation across edge, gateway, and cloud layers, the framework further enhances PdM effectiveness, ensuring that alerts are issued with context-appropriate responses and timing. Future work may further optimize the inference switching heuristics and explore additional edge device configurations to enhance adaptability in even more varied operational settings.

\section*{Acknowledgment}
The authors would like to thank ANID for its support through the scholarship ANID/Becas de Doctorado Nacional/2021-21210655.

% add bibliography.bib
\bibliographystyle{IEEEtran}  % Sets the bibliography style to IEEEtran
\bibliography{bibliography}   % Links to your bibliography.bib file

\begin{IEEEbiography}[{\includegraphics[width=1in,height=1.25in,clip,keepaspectratio]{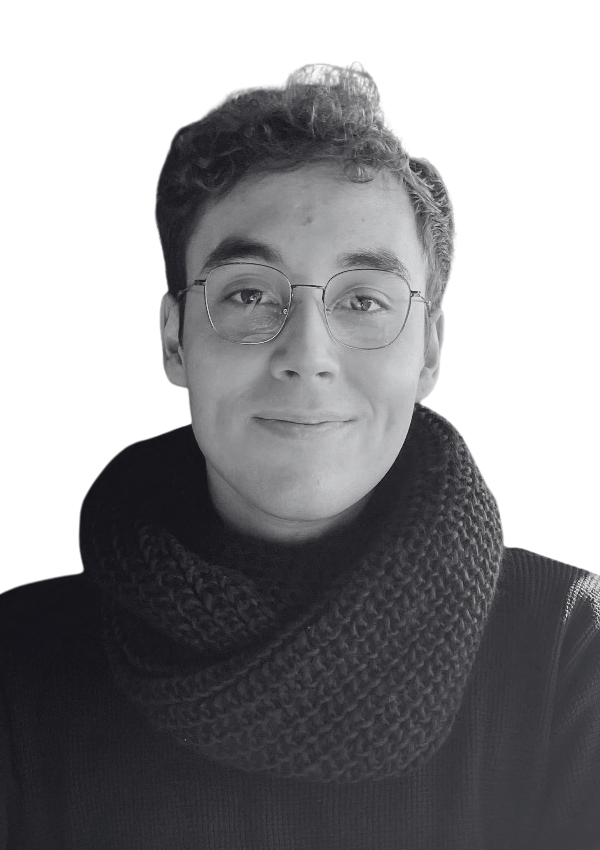}}]{Raúl de la Fuente} (Student Member, IEEE) was born in Santiago, Chile, in 1999. He received the B.Sc. degree in Computer Science from the University of Chile in 2021, graduating with maximum distinction. He is currently pursuing a combined Master's and Engineering degree in Computer Science and Engineering at the University of Chile, with an expected graduation date in December 2024. He serves as a Research Assistant in the Electrical Engineering Department at the University of Concepción, Chile, and at the Applied Cryptography and Cybersecurity Laboratory at the University of Chile. His research interests encompass embedded systems, IoT, computer architecture, hardware acceleration, applied cryptography, cybersecurity, and machine learning. 
\end{IEEEbiography}

\begin{IEEEbiography}[{\includegraphics[width=1in,height=1.25in,clip,keepaspectratio]{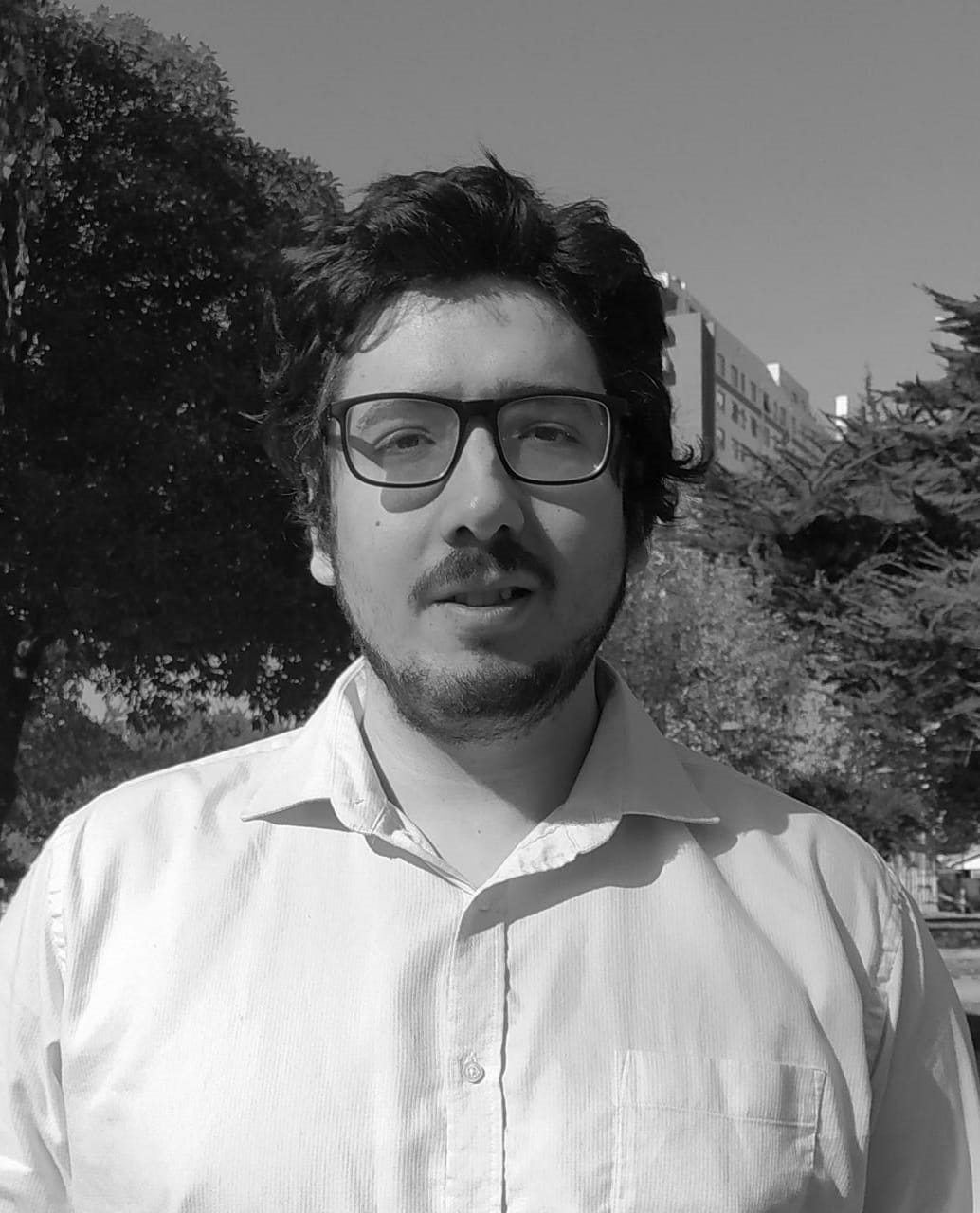}}]{Luciano Radrigan} was born in Chile. He received his B.S. in Electronic Engineering and M.S. in Electrical Engineering from the University of Concepción in 2019 and 2020, respectively. He is currently a Ph.D. candidate in Engineering, with a focus on Electrical Engineering, at the University of Concepción, Chile. In addition to his research, he serves as an adjunct professor in the Department of Electrical Engineering, Faculty of Engineering, at the University of Concepción. He is also an expert lecturer in the Department of Computer Science at the University of Chile, where he teaches elective courses on IoT and embedded systems. His research interests include condition monitoring sensors, IoT, industrial sensorization, machine learning, energy harvesting systems, embedded systems, and wireless energy transfer.

\end{IEEEbiography}

\newpage

%If you do not have or do not want to include a photo, you can use IEEEbiographynophoto as shown below:

\begin{IEEEbiography}[{\includegraphics[width=1in,height=1.25in,clip,keepaspectratio]{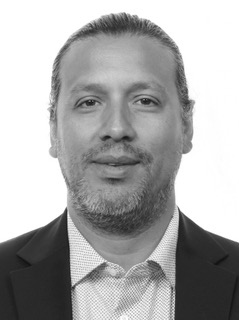}}]{Anibal S Morales} (Member, IEEE) was born in Concepcion, Chile, in 1983. He received the B.Sc. degree in Electronics engineering and the Ph.D. degree in Electrical engineering from the University of Concepcion, Chile, in 2007 and 2012, respectively. He joined University of Concepcion, Electrical engineering Dept. in 2012 as Postdoctoral research associate, followed by an Assistant professor position at Universidad Catolica de la Santisima Concepcion, Electrical engineering dept. from 2015 to 2024. He is currently an Assistant professor with Universidad San Sebastian, Chile since 2024. His current research interest include energy harvesting, energy efficiency \& electrical Safety, internet of things, sensors, cyberphysical systems \& technology for Industry Applications, copper electrowinning \& electrorefining, power electronics \& high-current rectifiers, and 3D Multiphysics Modeling with Finite Element Methods.

\end{IEEEbiography}

\EOD

\end{document}